\documentclass[letterpaper,twocolumn,10pt]{article}
\usepackage{usenix-2020-09}

\usepackage{tikz}
\usepackage{amsmath}

\usepackage{pifont}
\usepackage{algorithm}
\usepackage[noend]{algpseudocode}
\usepackage{subfig}
\usepackage{xspace}
\usepackage{enumitem}
\usepackage{makecell}
\usepackage{booktabs}

\newcommand{\parahead}[1]{\textbf{\textit{#1}}}
\def \SYS{FP6-LLM} 
\def \KERNEL{TC-FPx} 

\begin{document}

\date{}

\title{\Large \bf \SYS{}: Efficiently Serving Large Language Models Through FP6-Centric Algorithm-System Co-Design}

\author{
{\rm Haojun Xia}\\
University of Sydney
\and
{\rm Zhen Zheng}\\
Microsoft
\and
{\rm Xiaoxia Wu}\\
Microsoft
\and
{\rm Shiyang Chen}\\
Rutgers University
\and
{\rm Zhewei Yao}\\
Microsoft
\and
{\rm Stephen Youn}\\
Microsoft
\and
{\rm Arash Bakhtiari}\\
Microsoft
\and
{\rm Michael Wyatt}\\
Microsoft
\and
{\rm Donglin Zhuang}\\
University of Sydney
\and
{\rm Zhongzhu Zhou}\\
University of Sydney
\and
{\rm Olatunji Ruwase}\\
Microsoft
\and
{\rm Yuxiong He}\\
Microsoft
\and
{\rm Shuaiwen Leon Song}\\
Microsoft
}


\maketitle

\begin{abstract}
Six-bit quantization (FP6) can effectively reduce the size of large language models (LLMs) and preserve the model quality consistently across varied applications.
However, existing systems do not provide Tensor Core support for FP6 quantization and struggle to achieve practical performance improvements during LLM inference.
It is challenging to support FP6 quantization on GPUs due to (1) unfriendly memory access of model weights with irregular bit-width and (2) high runtime overhead of weight de-quantization.
To address these problems, we propose \KERNEL{}, the first full-stack GPU kernel design scheme with unified Tensor Core support of float-point weights for various quantization bit-width.
We integrate \KERNEL{} kernel into an existing inference system, providing new end-to-end support (called \SYS{}) for quantized LLM inference, where better trade-offs between inference cost and model quality are achieved.
Experiments show that \SYS{} enables the inference of LLaMA-70b using only a single GPU, achieving $1.69\times$-$2.65\times$ higher normalized
inference throughput than the FP16 baseline.
The source code is publicly available at \url{https://github.com/usyd-fsalab/fp6_llm}.
\end{abstract}

\section{Introduction}
Large Language Models (LLMs)~\cite{GPT3, GPT4, llama1, llama2, OPT-Models, AttentionIsAllYouNeed} are renowned for their capacity to process diverse language-related tasks~\cite{chen2021evaluating, Copilot, Bard, ChatGPT}.
However, it is challenging to deploy LLMs as these models are also characterized by their expansive size, e.g., 175 billion parameter GPT-3~\cite{GPT3} and 1.76 trillion parameter GPT-4~\cite{GPT4}.
On one hand, it requires large amounts of GPU memory (326 GB for GPT-3 in FP16) only to accommodate model weights, whereas an A100/H100 GPU~\cite{Ampere_WhitePaper, Hopper_WhitePaper} only has up to 80 GB memory.
On the other hand, LLM inference faces severe "memory wall" issues~\cite{FullStackOptimization, Flash-LLM} during token generation, where the speed of LLM inference is mainly limited by the time reading model weights from GPU DRAM.
It makes LLM inference \textit{memory bounded}, under-utilizing the computational power of GPUs.

Model quantization~\cite{zhu2023survey, OPTQ, awq, LLM.INT8, smoothquant, microscaling, ATOM} reduces both GPU memory footprint and DRAM data access.
It uses fewer bits to represent each model weight, resulting in a more compact representation of the model.
However, only a small set of bit-widths (i.e. 4-bit and 8-bit) are efficiently supported in existing systems~\cite{TensorRT-LLM, awq, awq-github, BitsAndBytes} on modern GPUs.
Recent studies show that 6-bit quantization is a good trade-off between inference cost and model quality for LLM deployment~\cite{zeroquant42, microscaling}.
However, there is still no efficient system support for the 6-bit linear layer execution (i.e., matrix multiplication) on modern GPUs.
It is urgent to develop the system support for 6-bit quantization fully leveraging the computing power of GPUs.

\begin{figure}
    \centering{
        \includegraphics[width=0.98\columnwidth]{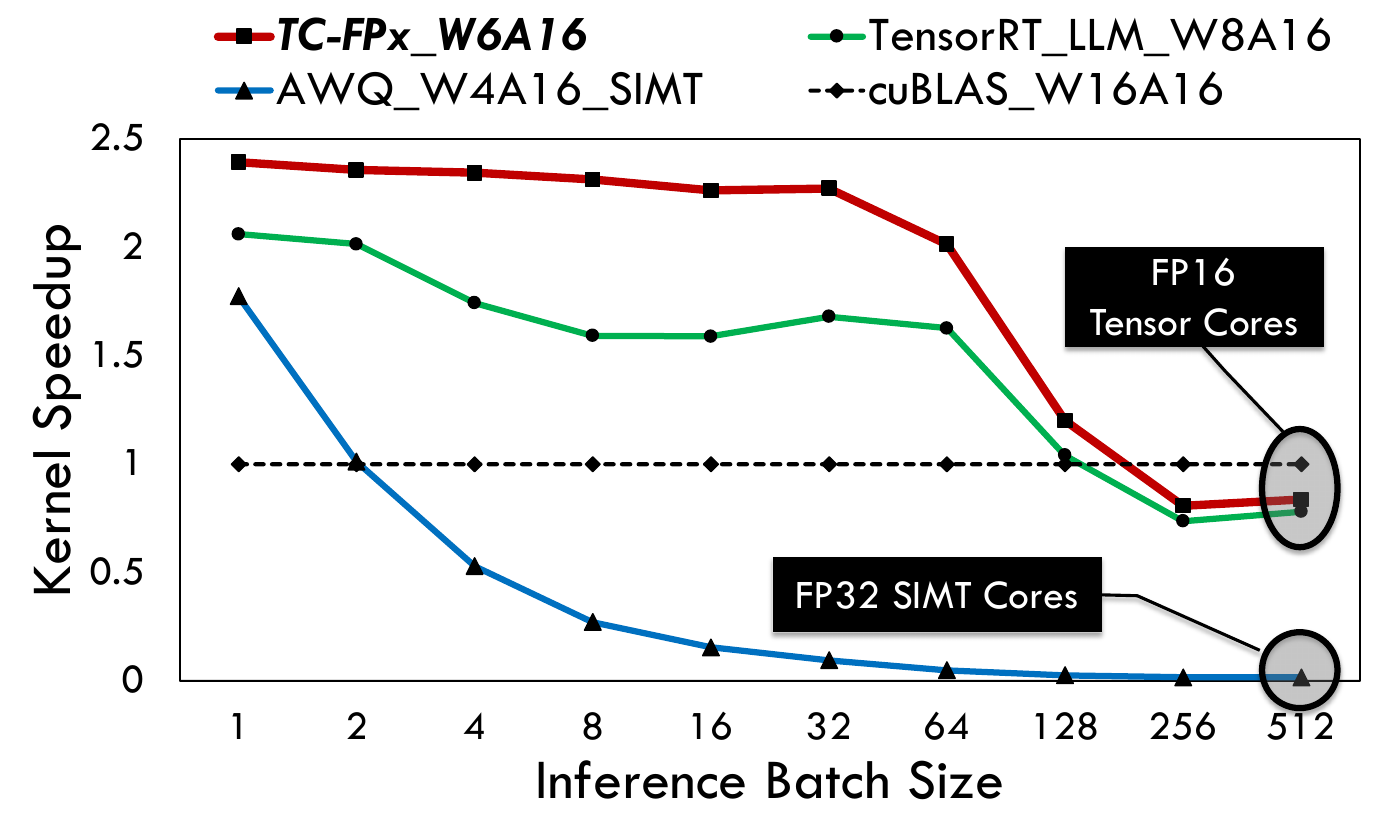}
    }
    \caption{Performance of a linear layer within the llama-65b~\cite{llama1} model. The shapes of the weight/activation matrices are (8192, 22016) and (22016, Batch Size).}
    \label{fig:ScalabilityOfSIMTKernles}
\end{figure}

On one hand, more efficient LLM inference can be achieved with 6-bit quantization compared to larger-bit quantization (e.g., 8-bit).
Firstly, more GPU memory can be saved, e.g. around 40 GB memory can be saved if deploying the GPT-3 model with 6-bit rather than 8-bit quantization.
Secondly, LLM inference can be further accelerated as the time of reading model weights from GPU DRAM can be effectively reduced.
As shown in Figure \ref{fig:ScalabilityOfSIMTKernles}, the linear layer implemented with our newly proposed 6-bit quantization system design (\KERNEL{}\_W6A16) is constantly faster (up to $1.45\times$) than the state-of-the-art support for 8-bit quantization (TensorRT\_LLM\_W8A16).
On the other hand, 6-bit quantization can more effectively preserve model quality compared to smaller-bit quantization (e.g., 4-bit). 
Despite the potential shown by recent 4-bit techniques~\cite{frantar2022gptq, yao2023zeroquant} in compressing LLMs with minimal quality loss, they are predominantly geared towards zero-shot evaluations. 
Recent research~\cite{zeroquant42} demonstrates that in tasks extending beyond zero-shot measurements, such as code generation and summarization, 4-bit methods underperform and lack robustness, whereas 6-bit quantization displays strong and consistent performance across these varied applications.

Motivated by the above observations, we propose \KERNEL{}, the first full-stack GPU system design scheme with unified Tensor Core~\cite{Ampere_WhitePaper, Hopper_WhitePaper} support of float-point weights for various quantization bit-width (6-bit, 5-bit, 3-bit, etc.), mitigating the "memory wall" issues during LLM inference.
\KERNEL{} breaks the limitations of the underlying GPU hardware, allowing the GPU to support linear layer calculations involving model weights of arbitrary bit width.
In \KERNEL{}, Tensor Cores are utilized for intensive computation of matrix multiplications, while SIMT cores are effectively leveraged for weight de-quantization, transforming the x-bit model weights to FP16 type during runtime before feeding them to Tensor Cores.
We propose \textit{Ahead-of-time Bit-level Pre-packing} (Section \ref{sec:WeightPrepacking}) to resolve the challenge of unfriendly memory access for weights with irregular bit-width (Section \ref{sec:DesignChallenge1}), enabling optimal GPU memory access.
Besides, we propose \textit{SIMT-Efficient GPU Runtime} (Section \ref{sec:GPURuntime}) to minimize the runtime overhead of weight de-quantization (Section \ref{sec:DesignChallenge2}).
Last but not least, we present the software pipeline of \KERNEL{} kernel, where SIMT cores, Tensor Cores, and the GPU memory hierarchy cooperate efficiently with high performance.

We integrate \KERNEL{} kernel into a state-of-the-art inference system~\cite{DeepSpeed}, providing new end-to-end support (called \SYS{}) for quantized LLM inference, where better trade-offs between inference cost and model quality are achieved.
Currently, \SYS{} mainly supports 6-bit quantization (FP6) for popular LLMs such as LLaMA~\cite{llama1}, OPT~\cite{OPT-Models} with various sizes.
Evaluations show that \SYS{} enables the inference of LLaMA-70b using only a single GPU, achieving $1.69\times$-$2.65\times$ higher normalized inference throughput than the FP16 baseline.
Besides, \SYS{} improves the inference throughput of OPT-30b by $1.72\times$-$4.05\times$.

\noindent In summary, we make the following contributions:
\begin{itemize}[leftmargin=*]
    \item We identify the significance and key challenges in supporting FP6 quantization on modern GPUs.
    \item We propose \KERNEL{}, the first full-stack GPU kernel design scheme with unified Tensor Core support of float-point weights with various bit-width, e.g. FP6.
    \item We provide new end-to-end inference support for quantized LLMs through the integration of \KERNEL{}, achieving better trade-offs between inference cost and model quality. 
    \item We evaluate \SYS{} on various LLM models and demonstrate that it substantially outperforms the baseline.
\end{itemize}
\section{Background}

\subsection{Quantization of Large Language Models}
\label{sec:Background_Quantization}
Although large language models (LLMs) are known for their impressive performance, their large size also creates challenges for model deployment.
Thus, model quantization~\cite{zhu2023survey, OPTQ, awq, LLM.INT8, smoothquant, microscaling, ATOM} is commonly used for LLM deployment, resulting in a more compact representation of the model. 
\textit{Weight-only quantization}~\cite{awq, OPTQ} only reduces the precision of model weights (e.g., INT8, using an 8-bit integer to represent each weight) while still using an FP16 value to represent each activation.
The major targets to be quantized are the weights of \textbf{linear layers} (i.e., matrix multiplication), which account for more than 99\% of the overall LLM weights.
The activations can also be quantized during inference~\cite{smoothquant, LLM.INT8}.
In this paper, we describe the precision of \textbf{W}eights and \textbf{A}ctivations with the term "\textbf{W}x\textbf{A}y", where $x/y$ denotes the bit-width of weights/activations.
Besides, the process of "dequantization" refers to transforming the quantized weights back to FP16.

\subsection{IEEE Standard for Floating-Point}
\label{sec:IEEE754}

The IEEE 754 float-point standard defines a binary format for representing real numbers.
Each floating point number consists of three parts: the sign bit (S), the exponent bits (E), and the mantissa bits (M). 
The corresponding value $f$ of a float-point number can be calculated via:
\begin{gather}
    f = (-1)^S \times (1.M) \times 2^{E-bias}; \quad bias = 2^{len(E)-1}-1
\end{gather}
Please refer to~\cite{IEEE754} for details, where special cases for values like infinity, zero, and NaN (Not a Number) are also defined.

\subsection{Tensor Cores vs. SIMT Cores}
\label{Sec:TC_SIMT}
\underline{SIMT cores} \footnote{Or referred to as CUDA cores on NVIDIA GPUs.} are responsible for \textbf{general-purpose} processing tasks in GPUs, which handle a wide range of instructions including integer operations, floating-point operations, load/store operations, etc.
SIMT cores execute scalar (or vector) instructions operating on individual (or vector) data elements.
\underline{Tensor cores}~\cite{Ampere_WhitePaper, Hopper_WhitePaper} are \textbf{specialized hardware} designed for accelerating matrix multiplication.
Tensor cores have $16.0\times$/$14.8\times$ higher FLOPS than SIMT cores on A100\cite{Ampere_WhitePaper}/H100~\cite{Hopper_WhitePaper} GPUs.
Besides, Tensor cores work at a coarse-grained granularity, e.g. performing a matrix multiplication between two FP16 matrices of shape 16 × 16 and 16 × 8 with a single \textit{mma} (matrix multiply and accumulate) instruction.
\section{Motivations}
\label{sec:BetterTradeoff}
8-bit~\cite{LLM.INT8, smoothquant} and 4-bit quantization~\cite{OPTQ, awq, ATOM} are the most widely applied schemes for the current post-training LLMs. 
However, recent algorithmic research~\cite{zeroquant42, microscaling} has demonstrated that superior trade-offs between inference cost and model quality can be achieved with FP6 quantization, compared to 8-bit and 4-bit quantization.

\parahead{(I) Lower inference cost than 8-bit quantization.}
Compared to the 8-bit quantization, the cost of deploying LLMs can be further reduced through more aggressive 6-bit quantization without a visible accuracy drop. 
On one hand, the size of LLM weights can be significantly reduced, nearly $2.7\times$ smaller than the FP16 baseline.
Less GPU memory is required to store model weights, thereby requiring fewer GPUs and reducing the serving cost of deploying LLMs.
On the other hand, 6-bit quantization can also more effectively accelerate the inference of LLMs.
Given that the LLM inference is usually \textit{memory-bounded}\footnote{When the execution is memory-bounded, it means that the rate at which data is transferred to or from the GPU's memory is the bottleneck, rather than the computational capabilities of the GPU cores.} during token generation, faster LLM inference can be achieved through reducing GPU DRAM access of the model weights.
As shown in Figure \ref{fig:ScalabilityOfSIMTKernles}, the execution of the linear layer within llama-65b model~\cite{llama1} is consistently faster (up to $1.42\times$ faster) with our newly proposed 6-bit quantization system design (\KERNEL{}\_W6A16) compared to the state-of-the-art 8-bit quantization support (TensorRT-LLM\_W8A16~\cite{TensorRT-LLM}).
Given that linear layers are the most time-consuming part of the large language models, this speedup will directly translate to performance improvements for end-to-end inference scenarios (See Section \ref{sec:e2e-inference}).

\parahead{(II) Better model quality than 4-bit quantization.}
Although 4-bit quantization more aggressively reduces memory footprint and DRAM access, it unavoidably causes degradation in model quality.
In contrast, near-lossless model compression can be achieved with 6-bit quantization.
As shown in Table \ref{tab:ZeroShot} and Table \ref{tab:GenerationTasks}, FP6 displays strong and consistent performance across various tasks including code generation and zero-shot perplexity performance. It also shows high robustness across various model sizes, e.g., 1B, 13B, and 65B LLaMA~\cite{llama1} models.
We also find that INT4 quantization heavily relies on \textit{Fine-Grained Quantization (FGQ)} methods to maintain high model quality, whereas our FP6 quantization already works well on coarse-grained quantization.
Note that the data points in Table \ref{tab:ZeroShot} and Table \ref{tab:GenerationTasks} are picked from~\cite{zeroquant42}. For more details, please refer to this paper.
In conclusion, at algorithmic level, FP6 quantization is a practical alternative to further democratize the deployment of LLMs without significantly sacrificing model quality on complex tasks and various model sizes. 

\begin{table}\footnotesize
\caption{Zero-shot evaluations, averaging over five datasets including PTB~\cite{marcinkiewicz1994building}, Wikitext~\cite{merity2016pointer}, and C4~\cite{raffel2020exploring}. Metric: perplexity, lower is better.} \label{tab:ZeroShot}
\begin{tabular}{lcccc}\toprule
                       & FP16  & FP6                        & INT4                       & INT4                       \\\midrule
Fine-Grain Quantization                   & /     & \ding{55}  &\ding{51}   & \ding{55} \\
LLaMA-1B   & 24.13 & 24.83                      & 564.73                     & 288.22                     \\
LLaMA-13B & 13.16 & 13.09                      & 14.19                      & 14.13                      \\
LLaMA-65B & 6.41  & 6.42                       & 6.61                       & 7.17  \\ 
\bottomrule
\end{tabular}
\end{table}

\begin{table}\footnotesize
\caption{Code Generation in HumanEval-X (JavaScript)~\cite{zheng2023codegeex}.} \label{tab:GenerationTasks}
\begin{tabular}{lcccc}\toprule
 &\multicolumn{3}{l}{}Metric: pass@1$\uparrow$, higher is better\\
         & FP16   & FP6                        & INT4                       & INT4                       \\\midrule
Fine-Grain Quantization     & /      & \ding{55} & \ding{51}   &\ding{55} \\
CodeGeeX2-6B~\cite{zheng2023codegeex}  & 31.50 & 31.61                      & 28.35                      & 25.15                      \\
 StarCoder-15B~\cite{li2023starcoder}& 33.67  & 33.6                       & 32.32                      & 32.18                      \\
CodeLLaMA-34B~\cite{mftcoder2023} & 45.05  & 44.51                      & 43.22                      & 43.45                 \\
\bottomrule
\end{tabular}
\end{table}

\section{Design Choices and Challenges}

\subsection{Design Choices}
\label{sec:choices}

Although there is an increasing demand for high-performance support of post-training FP6 quantization, currently there is no such efficient FP6-centric system design available that enables the aforementioned trade-offs against 4-bit and 8-bit quantization.   
Specifically, existing supports for linear layers are mainly designed for data types whose bit-width is an exponent of 2 (e.g., 4-bit, 8-bit, and 16-bit).
Given that it is not clear how to support FP6 efficiently on modern GPUs, we illustrate two important design choices in this section.

\parahead{Necessity in enabling Tensor Cores.}
We find it essential to support Tensor Cores when performing inference of quantized LLMs.
For example, we have evaluated the performance of AWQ's~\cite{awq-github, awq} pure SIMT-core execution on various batch sizes to test its scalability. 
As shown in Figure \ref{fig:ScalabilityOfSIMTKernles}, the runtime performance of linear layers without Tensor Core support (AWQ\_W4A16\_SIMT) becomes extremely low as the inference batch size increases.
The reason behind this is twofold.
On one hand, traditional SIMT cores are an order of magnitude slower than Tensor Cores for linear layer execution as described in Section \ref{Sec:TC_SIMT}.
On the other hand, a large fraction of the SIMT core's computational power will be used to de-quantize the model weights at runtime, which further reduces the available computational power of SIMT cores for computing matrix multiplication.
This motivates us to enable tensor cores for intensive computation of matrix multiplication while leveraging versatile SIMT cores for weight de-quantization.

\begin{figure}
    \centering
    \includegraphics[width=0.9\columnwidth]{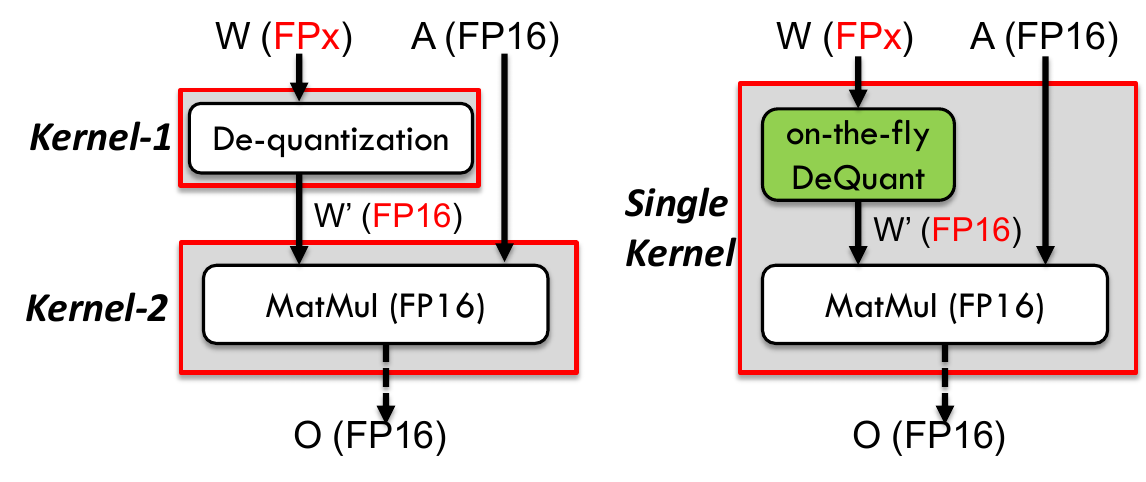}
    \caption{Two different methods to support weight-only WxA16 quantization during LLM inference. (Left) Dual kernels. (Right) Unified kernel.  
    }
    \label{fig:SingeKernel_vs_TwoKernels}
\end{figure}

\parahead{Unified kernel solution rather than dual kernels.}
The unique character of \textit{WxA16 quantization} is that the activation matrices use FP16 but the weight matrices are stored in a narrower bit-width.
However, Tensor Cores require both the weights and activations matrices to be stored in the same data type, e.g. FP16/INT8/INT4.
The straightforward solution (i.e., dual kernel solution) adds an extra GPU kernel that de-quantizes the weights to FP16 before calling the normal FP16 kernel.
However, such inference speed would be even slower than that of the model without quantization.
As shown in Figure \ref{fig:SingeKernel_vs_TwoKernels} (Left), two GPU kernels will be launched for the linear layer execution, and the de-quantized FP16 weights will be written to GPU DRAM before being read by the second GPU kernel, resulting in $2\times$ DRAM access.
It is more efficient to fuse the de-quantization and the matrix-multiply process into a single GPU kernel, eliminating the read/write of the de-quantized weights ($W^{'}$ in FP16).

\subsection{Design Challenges}
\label{sec:DesignChallenges}

Given the design choices in Section \ref{sec:choices}, it is challenging to design a unified GPU kernel supporting FP6$\times$FP16 matrix multiplication on modern GPUs.
On one hand, modern GPU memory systems do not naturally support irregular bit-width (not an exponent of 2) because the minimal access size of GPU global/shared memory is 8/32 bits per thread and the memory addresses to access must be aligned.
The complex data layout requirement of Tensor Cores makes it even more challenging for irregular bit-widths.
On the other hand, the de-quantization computation is expensive as it requires a large amount of complex bit-level operations.
Thus, how to fuse the de-quantization into the linear layer computation without hurting the overall performance is also non-trivial.

\subsubsection{Hardware-Unfriendly Memory Access}
\label{sec:DesignChallenge1}
During the execution of linear layers on modern GPUs, model weights should be loaded from DRAM to \textit{registers} before the corresponding multiplication calculations can be performed.
Usually, the model weights are loaded in two steps, to hide the high access latency of DRAM for high performance.
Specifically, model weights are first loaded from GPU DRAM and buffered into on-chip memory (e.g., \textit{shared memory}) for data reusing.
After that, the buffered weights are then read from \textit{shared memory} to \textit{registers} for the actual computation. 

\begin{figure}
    \centering
    
    \subfloat[Required Data Layout of Tensor Cores Input. T0 Means Thread \#0.
    \label{fig:DataLayout_TC}]{
        \includegraphics[width=0.98\columnwidth]{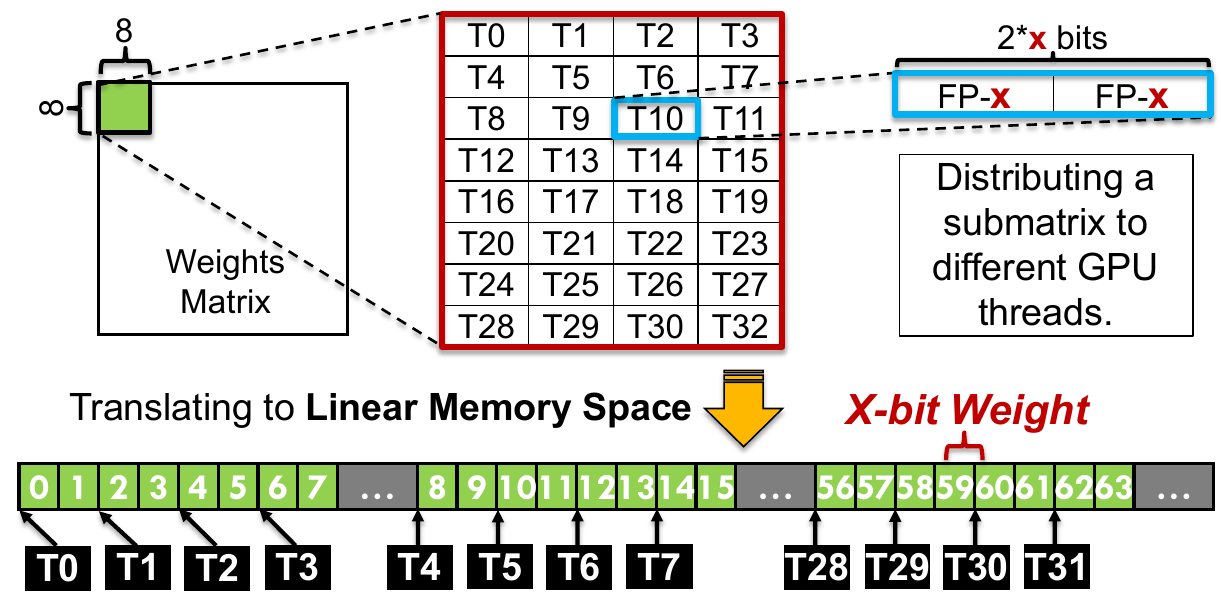}
    }
    
    \subfloat[Accessing 6-bit weights at the granularity of 32-bit Words.
    \label{fig:IrregularBitWidth}]{
        \includegraphics[width=0.98\columnwidth]{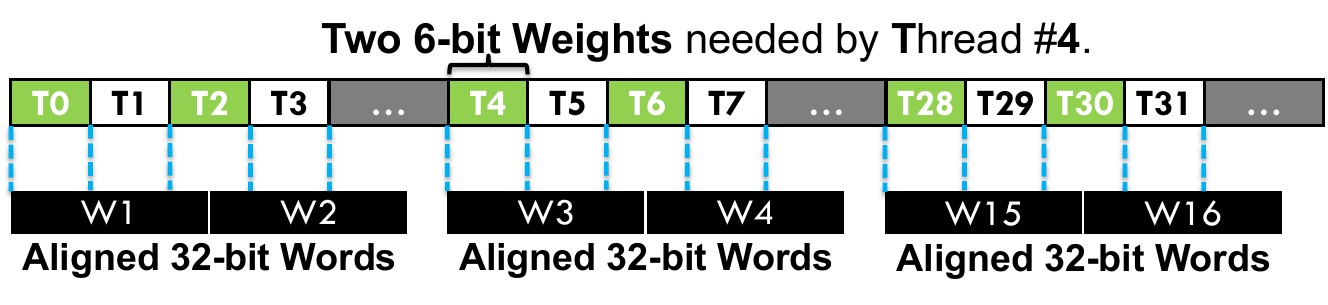}
    }
    \caption{Memory Access of X-bit Weights for each Thread.}
    \label{fig:}
\end{figure}

Given that each GPU thread \textbf{cannot} directly access other threads' \textit{registers}\footnote{Each GPU thread is allocated and owns a distinct portion of the whole registers available on GPU processors.}, each thread must put the model weights that are needed by itself to its private \textit{registers} \textbf{on its own}. 
This process can become extremely challenging when the weights are stored with irregular bit-width (not $2^n$, e.g., 6 bit), given the rigid data layout requirements of Tensor Cores.
As shown in Figure \ref{fig:DataLayout_TC}, the minimal input of FP16 Tensor Cores is a $8\times8$ sub-matrix in modern GPU architecture, and each GPU thread should hold a pair of weights in its \textit{register}.
In normal cases, each weight is stored with 16 bits, and each pair of weights can be naturally read from \textit{shared memory} at the granularity of 32-bit words.
However, each weight is stored with x-bits in our work\footnote{Our design principles support not only 6-bit but also any other bit widths.}, which makes memory access extremely unfriendly to modern GPU memory hierarchy.

\underline{On-chip Memory Access with Unused Bits:}
We use 6-bit quantization as an example to show the inefficiency in accessing weights with irregular bit-width.
As shown in Figure \ref{fig:IrregularBitWidth}, weights are already buffered in \textit{shared memory}, and each GPU thread needs to read a pair of weights (\textbf{12 bits}, $2*6 bits$) from \textit{shared memory}.
However, \textit{shared memory} has 32 memory banks and each memory bank outputs a \textbf{32-bit} word per memory request on GPU.
Thus, a large fraction of bits read from shared memory will be unused, resulting in a significant waste of shared memory bandwidth.
For instance, T0 (Thread \#0) in Figure \ref{fig:IrregularBitWidth} only needs 12 bits.
However, a 32-bit word (W1) will be read, resulting in 20 out of 32 bits ($62.5\%$) unused.
The waste of unused bits can get even more severe due to the requirement of \textbf{aligned memory access}\footnote{Memory access must be aligned, i.e., its address is a multiple of its size.} in modern GPU memory hierarchy.
As shown in Figure \ref{fig:IrregularBitWidth}, the bits needed by T2 (Thread \#2) are distributed in both W1 and W2.
Thus, T2 needs to read both W1 and W2, reading $2*32$ bits from \textit{shared memory}.
However, only $6*2$ bits will be eventually used, resulting in 52 out of 64 bits ($81.25\%$) unused and wasted.
It is also worth noting that the memory management and access on GPU DRAM and \textit{registers} suffer from similar problems due to the irregular bit-width.

\subsubsection{High Computation Overhead of De-quantization}
\label{sec:DesignChallenge2}
The runtime overhead of FPx-FP16 de-quantization can be extremely high, which easily slows down the overall execution.
On one hand, large amounts of model weights need to be de-quantized at runtime, e.g. 70 billion FPx weights should be de-quantized for each LLM decoding step\footnote{To generate a sequence with n tokens, n-1 decoding steps are required.} for LLaMA-70b~\cite{llama2} inference.
On the other hand, the runtime overhead to de-quantize each FPx weight is high, requiring complex bit-wise operations.
According to Equation \ref{equ:new_exp}, new sign, exponent, and mantissa all need to be calculated during runtime, to obtain the FP16 with the equivalent value of a given FPx. 
\begin{equation} \label{equ:new_exp}
    2^{E^{fp16} - bias^{fp16}} \times (1.M^{fp16}) =  2^{E^{fpx} - bias^{fpx}} \times (1.M^{fpx})
\end{equation}
In Equation \ref{equ:new_exp}, $bias^{fp16} = 15$ and $bias^{fpx} = 2^{len(E^{fpx})-1}-1$.
The sign field of the FP16 is identical to that of the FPx, and the mantissa of the FP16 can also be calculated by padding zeros to that of the FPx.
What's more, the exponent of FP16 should be $E^{fp16} = E^{fpx} + bias^{fp16}-bias^{fpx}$, which is more computationally expensive.
In summary, how to de-quantize FPx values efficiently also becomes a major challenge.

\section{Design Methodology}

In this section, we first provide an overview of our designs in Section \ref{sec:DesignOverview}.
To solve the challenge of unfriendly memory access (Section \ref{sec:DesignChallenge1}), we propose \textit{Ahead-of-time Bit-level Pre-packing} in Section \ref{sec:WeightPrepacking}.
To deal with the challenge of the high computational overhead of de-quantization (Section \ref{sec:DesignChallenge2}), we presented our designs to achieve \textit{SIMT-Efficient GPU Runtime} in Section \ref{sec:GPURuntime}.
At last, we presented our software pipeline designs in Section \ref{sec:PipelineDesign}, where SIMT cores, Tensor Cores, and GPU memory hierarchy work collaboratively with full performance.

\subsection{Overview}
\label{sec:DesignOverview}
\begin{figure}
    \centering
    \subfloat[\KERNEL{} Design.
    \label{fig:DesignOverview1}]{
        \includegraphics[width=0.44\linewidth]{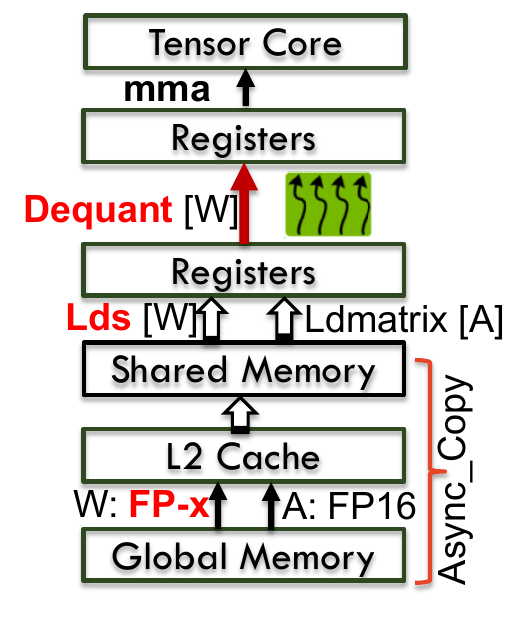}
    }
    \subfloat[Traditional GEMM Design.
    \label{fig:DesignOverview2}]{
        \includegraphics[width=0.4\linewidth]{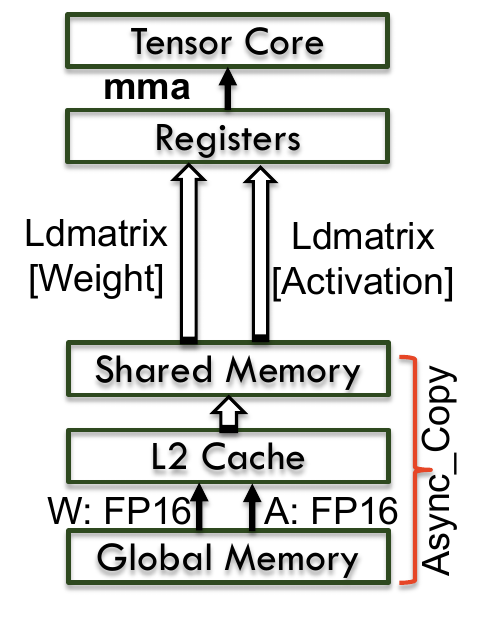}
    }
    \caption{Design Overview.}
    \label{fig:DesignOverview}
\end{figure}

Figure \ref{fig:DesignOverview} compares \KERNEL{}, the x-bit weight-only quantized linear layer kernel in our design, with the traditional design for general-purpose matrix multiplication (GEMM) where both input matrices are in FP16.
The model weight is stored with a reduced number of bits for \KERNEL{}.
Consequently, an additional de-quantization stage (Dequant W) is introduced at the register level, where the FP6 weights are de-quantized to FP16 locally within each thread using SIMT cores.
It is worth noting that the FP16 weights are not written back to \textit{shared memory} but stored in \textit{registers} for future use, eliminating unnecessary round-trip access to \textit{shared memory}.
Another difference is that \KERNEL{} loads x-bit weights from \textit{shared memory} to \textit{registers} using fine-grained \texttt{lds} (load shared) instructions instead of using the coarse-grained intrinsic \texttt{ldmatrix} (load matrix), which has a strict layout requirement and is less flexible.

\subsection{Ahead-of-time Bit-level Pre-packing}
\label{sec:WeightPrepacking}

As described in Section \ref{sec:DesignChallenge1}, memory access to weights with irregular bit-width is unfriendly to modern GPU memory hierarchy.
To address this problem, we propose the insight that we can combine the memory read of every \textbf{32 x-bit weights}, resulting in \textbf{x request of 4-byte word} per GPU thread.
In this case, all the memory access would be aligned at the granularity of 32-bit words rather than the irregular bit-width.

However, it is not trivial to combine the memory read of weights due to the rigid data layout requirements of Tensor Cores, because the weights needed by each GPU thread are not stored in continuous memory space.
To solve this problem, we propose to optimize the runtime memory access pattern by reordering the weights within each weight matrix and pre-pack the weights ahead of time.
As model weights are statically determined after the model is trained and quantized, complicated memory layout transformation can be applied to the weights ahead of time and thus introduces no runtime overhead.
Besides, we only need to pre-pack the weights once, thus the overhead of weight pre-packing can be effectively amortized by each inference service and becomes negligible.

In general, weight pre-packing consists of two steps.
In the first step, we gather all the weights needed by each GPU thread and combine these weights locally.
Given that the weights needed by each GPU thread are not originally in continuous locations (see Figure \ref{fig:DataLayout_TC}) within each weight matrix, we must pick the weights for each GPU thread carefully.
The weights picked for each thread are then combined locally in relative temporal order as they are consumed by Tensor Cores at runtime.
In the second step, we combine all the weights needed by the whole GPU WARP (consisting of 32 GPU threads) into a unified linear memory space, in which order the weights will be stored in GPU DRAM before runtime.
To fully eliminate \textit{shared memory} bank conflict\footnote{Bank conflicts occur in shared memory when multiple threads access data in the same memory bank simultaneously, leading to lower throughput.}, we propose to combine the 32-bit word of each thread in a jagged order.

It is worth noting that all the techniques discussed in this subsection are independent of the actual bit-width (denoted using x the whole time) of the model weights.
Thus, our weight pre-packing can be naturally applied to any bit-width.

\begin{figure}
    \centering
    \includegraphics[width=0.9\columnwidth]{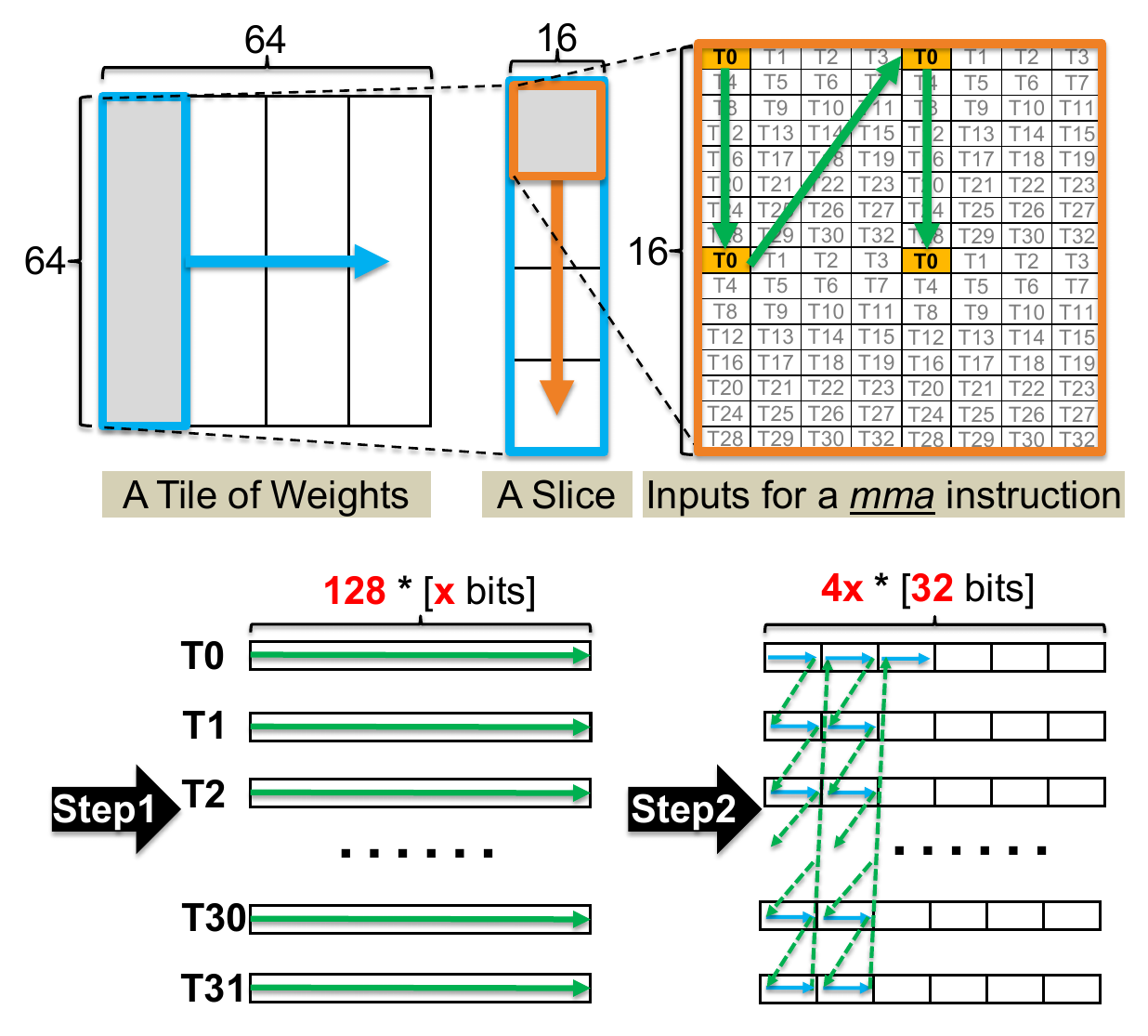}
    \caption{Ahead-of-time Bit-level Weight Pre-packing.
    }
    \label{fig:WeightRearragement}
\end{figure}

\paragraph{Step 1: Per-thread Weight Gathering}
Figure \ref{fig:WeightRearragement} demonstrates the weights picked by T0 (Thread \#0) and the order to combine them.
We suppose the WARP-level tiling size is $64\times64$, which means each weight matrix is divided into $64\times64$ data tiles and loaded to GPU's \textit{shared memory} at this granularity for each WARP.
Each weight tile is then further divided into four slices, as the weights are loaded from \textit{shared memory} and used for Tensor Core computations on a slice-by-slice basis.
What's more, each slice is divided into four $16\times16$ chunks, as Tensor Core processes $16\times16$ data items in each instruction.
Within each $16\times16$ chunk, four pairs of FPx weights are picked for T0 and combined.
As shown in Figure \ref{fig:WeightRearragement}, we get 32 (i.e., the WARP size) groups of FPx weights after Step 1.
The weights are combined and stored continuously within each group and
each group of weights will be consumed by a certain GPU thread.
In summary, each $64\times64$ weight tile is eventually assigned to 32 threads (a WARP), and each thread will consume 128 x-bit weights.

\paragraph{Step 2: Bit-level Assembling per WARP}
In Step 2, we assemble all the weights of different groups into a unified memory space.
During this bit-level pre-packing process, we consider the combined weights as continuous data to copy, temporarily ignoring the meaning of each bit.
Specifically, 128 items with x-bit are considered as 4x items with 32 bits.

We propose to assemble the weights of all groups in the \textbf{jagged order} shown in Figure \ref{fig:WeightRearragement}.
To begin with, the first 32-bit item of each thread is concatenated together.
After that, the second 32-bit item of each thread is concatenated and appended to the previous results.
By repeating this process, all weights can be stored continuously in a linear memory space and well-aligned (128-byte aligned).
In this way, all weights can be simply copied from DRAM to \textit{shared memory} at the granularity of 128-byte blocks without any changes, easily achieving optimal DRAM access.
Besides, these weights can then be loaded from \textit{shared memory} with optimal performance as well during runtime.
Specifically, a WARP of threads will read consecutive 32-bit items in \textit{shared memory} for each memory request, fully avoiding bank conflict.

\subsection{SIMT-Efficient GPU Runtime}
\label{sec:GPURuntime}

\begin{figure*}
    \centering
    \subfloat[Optimized FP6 to FP16 cast.
    \label{fig:SimplifiedDequant}]{
        \includegraphics[width=0.25\linewidth]{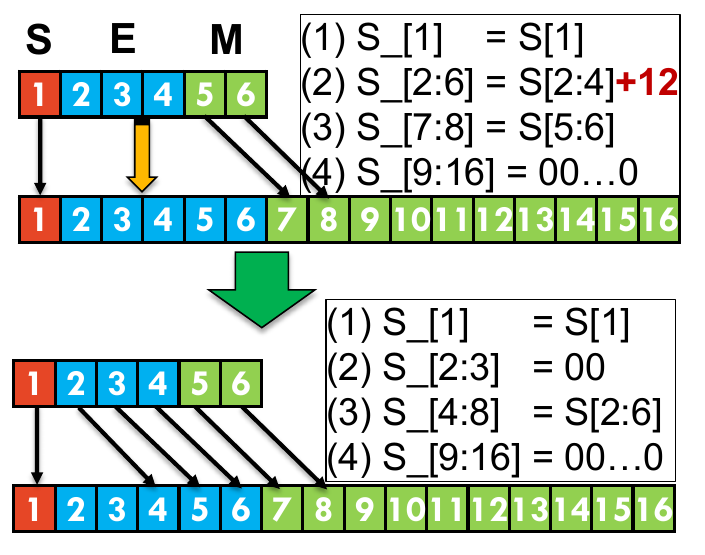}
    }
    \subfloat[4-Way parallel de-quantization within 32-bit registers.
    \label{fig:ParallelDequant}]{
        \includegraphics[width=0.65\linewidth]{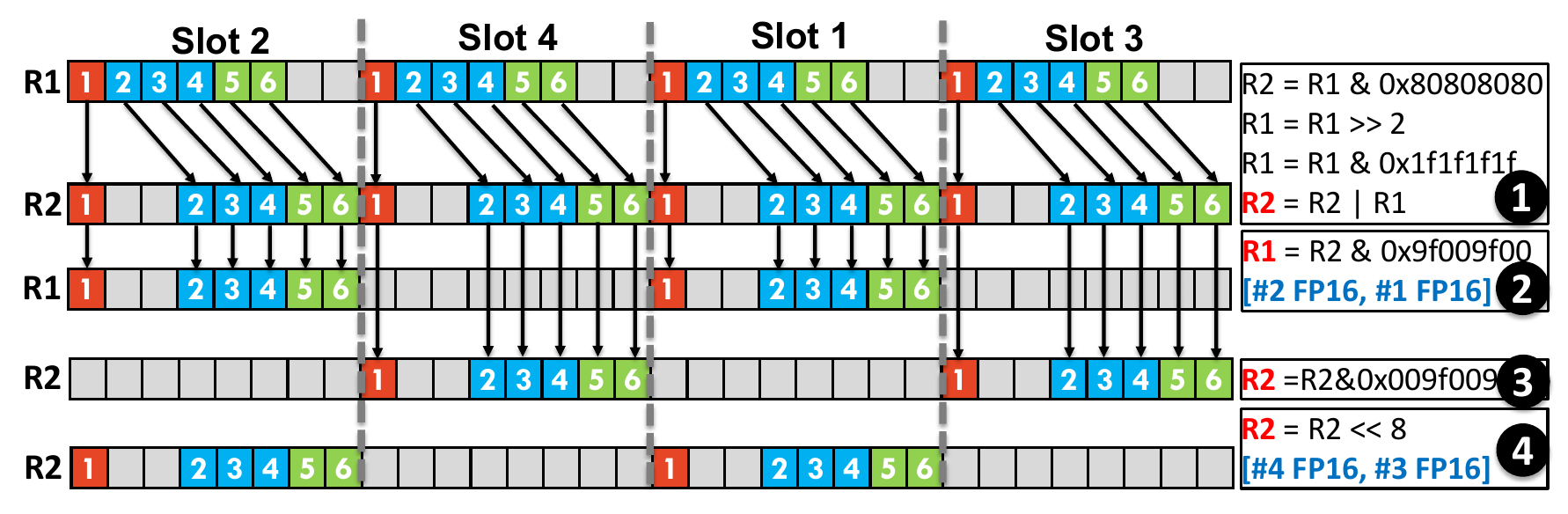}
    }
    \caption{SIMT-Efficient On-the-fly Parallel De-quantization.}
    \label{fig:SIMTEfficientDequant}
\end{figure*}

\paragraph{Parallel De-quantization}
To reduce the runtime overhead of FP-x weight de-quantization, we implemented FP-x de-quantization with optimized bit-wise SIMT core instructions.
Besides, we propose to de-quantize multiple FPx weights in parallel, further reducing the SIMT overhead by $4\times$ by exploiting the bit-level parallelism within each 32-bit register.

\underline{(1) Optimized Bit-wise Operations:}
As described in Section \ref{sec:DesignChallenge2}, the exponent for FP16 should be $E^{fp16} = E^{fpx} + bias^{fp16}-bias^{fpx}$, when casting an FPx to the equivalent FP16.
To simplify this process, we adopted the mathematical transformation in~\cite{zeroquant42}, calculating the exponent of FP16 with $E^{fp16} = E^{fpx}$ instead.
To maintain correctness, the result FP16 is then multiplied with the FP16 constant $2^{bias^{fp16}-bias^{fpx}}$:
\begin{equation} \label{equ:FPx-cast}
    cast(W_{fpx}) = new\_cast(W_{fpx}) \times 2^{bias^{fp16}-bias^{fpx}}.
\end{equation}
Fig.\ref{fig:SimplifiedDequant} shows the optimized FP16 to FP6 conversion.
Although we only draw the cast from FP6 to FP16 for demonstration, it can be applied to any bit-width.
The sign field of FP16 is identical to that of FPx.
Besides, the lower bits of the exponent field and the higher bits of the mantissa field can be copied from FPx to FP16 together for efficiency.
What's more, other bits of FP16 should be padded with zeros.

With careful designs, we succeeded in achieving cast from FP6 to FP16 with only two bit-wise \textit{"and"}, one \textit{"shifting"}, and one \textit{"or"} as shown in \ding{182} of Figure \ref{fig:ParallelDequant}.
The sign field is copied from FP6 to FP16 with the first \textit{"and"} and all other bits of the FP16 are initialized to zeros at the same time eliminating the need to pad zeros to the exponent and mantissa fields later.
All bits of the FP6 are then shifted right with the bit-wise \textit{"right shifting"}.
After that, the lower bits of the exponent and the higher bits of the mantissa in FP6 are first selected via the \textit{"and"} between the FP6 and the bit mask "0x1f1f1f1f", and then copied to the FP16 with the bit-wise operation \textit{"or"}.

\underline{(2) Bit-level Parallelism:}
Given the insight that we can exploit the bit-level parallelism within each 32-bit word, we propose to de-quantize multiple FPx weights in parallel, further reducing the runtime overhead of de-quantization.
The detailed design is demonstrated in Figure \ref{fig:ParallelDequant} using FP6 as an example.
The 32-bit registers are treated as four processing slots, where each slot works independently with the same instruction but different input FP6 data.
Before the start of de-quantization, four FP6 should be stored in $R1$ (Register \#1) with the initial data layout shown in the figure.
With the code snippet \ding{182}, these four FP6 can be simultaneously de-quantized into four FP16, where only the first 8 bits of each FP16 are stored in $R2$.
After that, the first and the second FP16 are extracted to $R1$ with their last 8 bits padded with zeros, with the code snippet \ding{183}.
Finally, with the code snippet \ding{184} and \ding{185}, the third and the fourth FP16 are extracted to $R2$.

\paragraph{Weight Split and Stitching}
We will then demonstrate the method to efficiently reconstruct the 6-bit weights from the 2+4 scheme~\cite{zeroquant42} on GPUs with a carefully designed memory layout, which can also be applied to other bit-width.

\underline{(1) Ahead-of-time Weight Split:}
To store the weights in a well-aligned manner in GPU's 32-bit \textit{registers},
we split each weight into several segments, where the bit-width of each segment is $2^{n}$, e.g. each 6-bit weight can be split into either 2+4 or 4+2.
Based on this scheme, the index calculations for the following designs are significantly simplified.
Note that the techniques described in Section \ref{sec:WeightPrepacking} can be applied to any bit-width, thus the 2-bit and 4-bit segments can be pre-packed separately and efficiently according to Section \ref{sec:WeightPrepacking}.

\begin{figure}
    \centering
    \includegraphics[width=0.9\columnwidth]{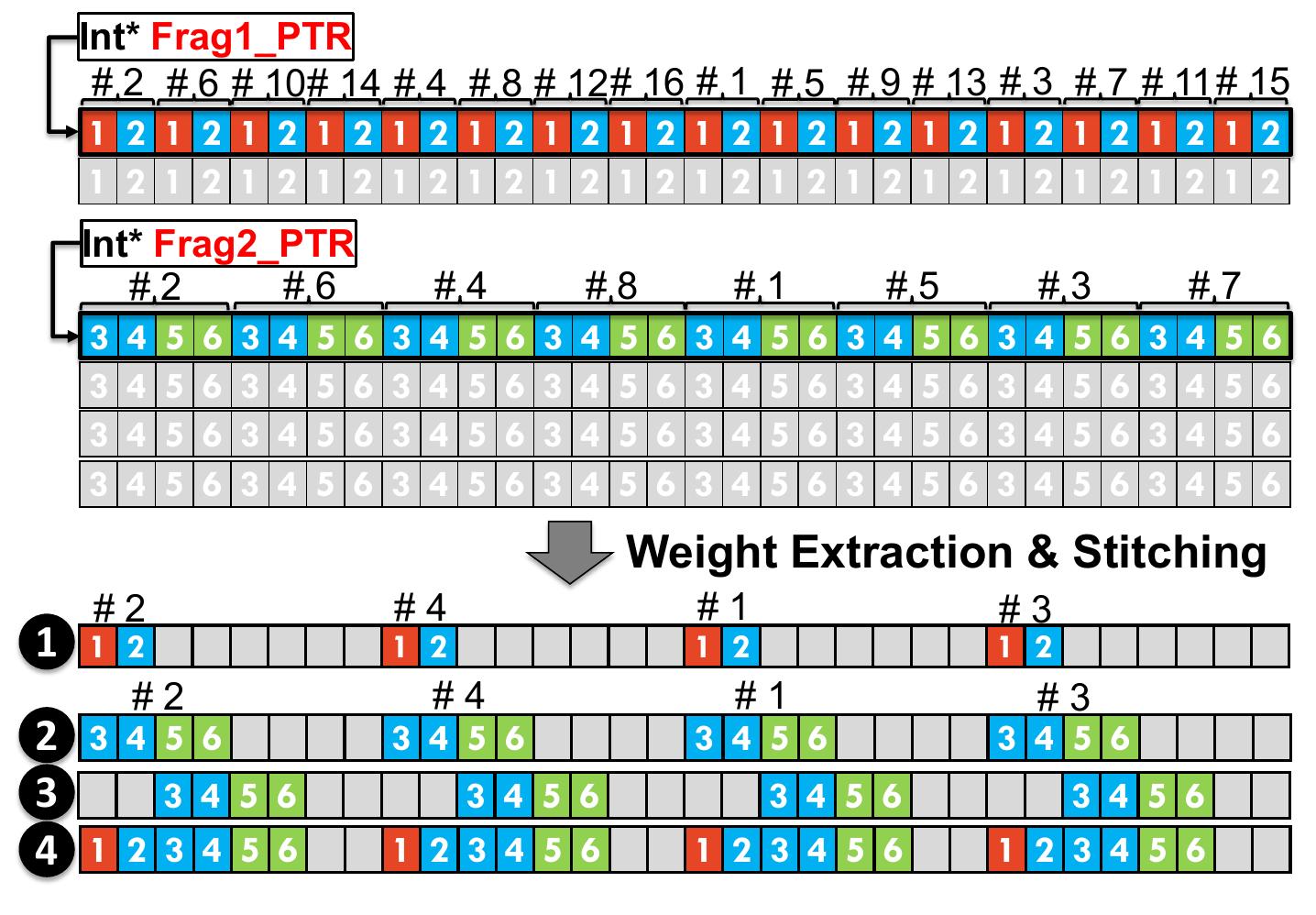}
    \caption{Parallel Weight Stitching.}
    \label{fig:WeightStitching}
\end{figure}

\underline{(2) Runtime Weight Stitching:}
Before the de-quantization, the weights are first loaded from \textit{shared memory} to \textit{registers}.
As each weight is split into several segments, the complete weights need to be reconstructed at the register level during runtime.
To reduce this runtime overhead, we propose to extract and stitch the weights in parallel.
As shown in Figure \ref{fig:WeightStitching}, two sets of registers are used to store 32 FP6 weights, where $Frag1\_PTR$ points to two 32-bit registers containing 32 2-bit segments while $Frag2\_PTR$ points to four 32-bit registers containing 32 4-bit segments.
With our parallel stitching, four FP6 weights are reconstructed simultaneously, reducing the number of SIMT core instructions by $4\times$.
As shown in Figure \ref{fig:WeightStitching}, four 2-bit segments are first extracted to Register \#1 (\ding{182}), and four 4-bit segments are then extracted to Register \#2 (\ding{183}).
After that, Register \#2 is right-shifted (\ding{184}) and its valid bits are copied to Register \#1 (\ding{185}), resulting in complete 6-bit weights.

\underline{(3) Bit Reordering:}
To extract and stitch the weight in parallel, it is necessary to enforce the initial data layout in Figure \ref{fig:WeightStitching}.
The key observation is that each four continuous segments must be placed in the order shown in the figure, e.g. the first four segments must be stored in the order of \#2, \#4, \#1, and \#3.
Besides, the stride between each pair of 2/4-bit segments should be 6/4, respectively.
Otherwise, it is not possible to stitch four segments simultaneously with only four SIMT core instructions.
To satisfy the initial data layout requirements in Figure \ref{fig:WeightStitching}, we propose to ensure this layout via reordering the weight segments before runtime with no runtime overhead.
Besides, this technique is supposed to be superimposed on the technique described in Section \ref{sec:WeightPrepacking} as an additional pass.

\paragraph{Overall Pseudo Code}
Algorithm \ref{alg:WeightStitching} shows the pseudo code (GPU code) including both \textit{Parallel De-quantization} and \textit{Weight Stitching}.
All the input and output variables in the pseudo code are stored in \textit{registers}.
As demonstrated in Figure \ref{fig:WeightStitching}, Algorithm \ref{alg:WeightStitching} de-quantizes 32 FP6 weights in total.
For each outer loop, four FP16 weights are generated and stored with two \textit{registers} at the end of the code.
The transformations in Figure \ref{fig:WeightStitching} (\ding{182}, \ding{183}, \ding{184}, and \ding{185}) are achieved with the SIMT core operations of lines 6, 7, 9, and 10 in Algorithm \ref{alg:WeightStitching}, respectively.
The output \textit{register} array (\textit{OutputReg}) is then directly used by Tensor Cores as inputs.

\begin{algorithm}\footnotesize
\caption{Weight Stitching \& De-quantization.} 
\label{alg:WeightStitching}
\begin{algorithmic}[1]
\State \textbf{Inputs:} \text{\textcolor{blue}{int} $Frag1\_ptr[]$, \textcolor{blue}{int} $Frag2\_ptr[]$, \textcolor{blue}{half} $Scales[]$}
\State \textbf{Output:} \text{\textcolor{blue}{int} $OutputReg[]$}

\State \textcolor{red}{\#pragma} \text{unroll}
\For {\textcolor{blue}{int} $i = 0;$ $i<8;$ $i++$}
    \State \textcolor{brown}{//Weight Extraction}
    \State \textcolor{blue}{unsigned int} $R1 = (*Frag1\_ptr) \& 0xc0c0c0c0;$ \Comment{\ding{182}}
    \State \textcolor{blue}{unsigned int} $R2 = (*Frag2\_ptr) \& 0xf0f0f0f0;$
    \Comment{\ding{183}}
    \State \textcolor{brown}{//Weight Stitching}
    \State $R2 = R2 >> 2;$
    \Comment{\ding{184}}
    \State $R1 = R1 | R2;$
    \Comment{\ding{185}}
    \State \textcolor{brown}{//Advancing to next register or shifting current register.}
    \If{$i\%4==3$}
        \State $Frag1\_PTR++;$
    \Else
        \State $(*Frag1\_PTR) = (*Frag1\_PTR) << 2;$
    \EndIf
    \If{$i\%2==1$}
        \State $Frag2\_PTR++;$
    \Else
        \State $(*Frag2\_PTR) = (*Frag2\_PTR) << 4;$
    \EndIf    
    \State \textcolor{brown}{//4-Way Parallel de-quantization.}
    \State $*R2 = *R1 \& 0x80808080; $
    \State $*R1 = *R1 >> 2; $
    \State $*R1 = *R1 \& 0x1f1f1f1f; $
    \State $*R2 = *R2 | *R1; $
    \State $*R1 = *R2 \& 0x9f009f00; $
    \State $*R2 = *R2 \& 0x009f009f; $
    \State $*R2 = *R2 << 8; $ \Comment{\textcolor{teal}{R1 and R2 now each has 2 FP16 weights.}}
    \State \textcolor{brown}{//Multiplying with quantization scales \& Output to registers.}
    \State $OutputReg[i*2]=Multiply(R1, Scales[i/2*2]);$
    \State $OutputReg[i*2+1]=Multiply(R1, Scales[i/2*2+1]);$
\EndFor
\end{algorithmic}
\end{algorithm}

\subsection{Software Pipeline Design}
\label{sec:PipelineDesign}

To reduce the usage of GPU registers, we de-quantize the weights slice by slice.
Besides, we seamlessly fuse the process of de-quantization into the traditional software pipeline of linear layer execution, completely hiding the runtime overhead of de-quantization via effective instruction parallelism.

\begin{figure}
    \centering
    \subfloat[Slice-by-slice De-quantization. 
    \label{fig:FourStageDequant}]{
        \includegraphics[width=0.75\linewidth]{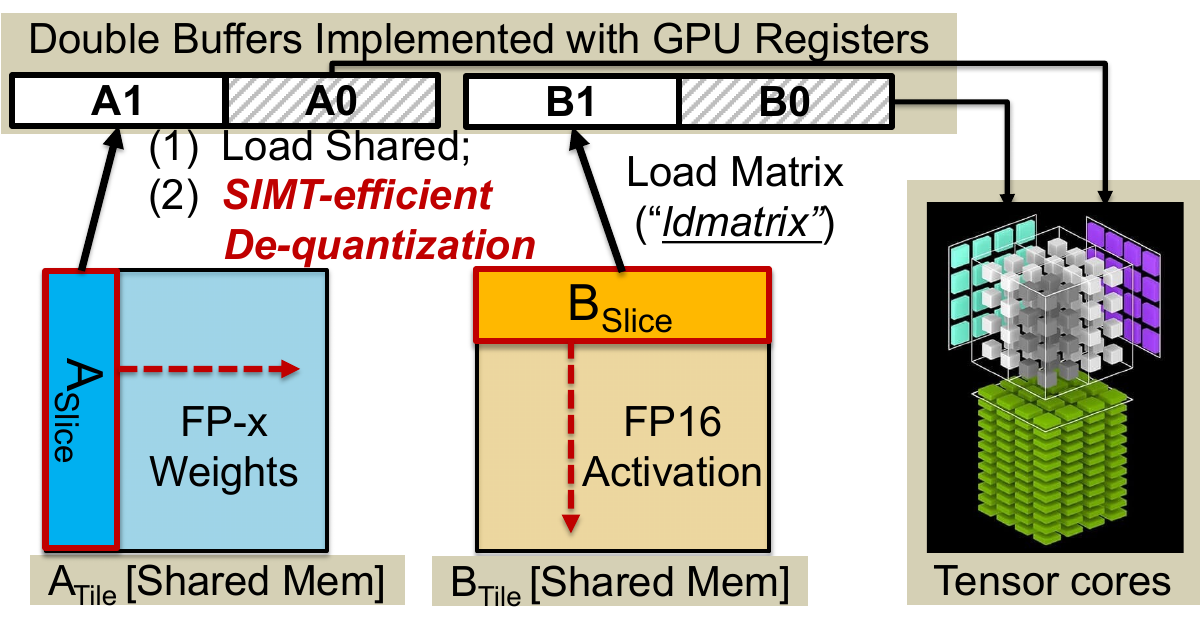}
    }
    
    \subfloat[Space-time Diagram of the Kernel Pipeline.
    \label{fig:SpaceTimeDiagram}]{
        \includegraphics[width=0.98\linewidth]{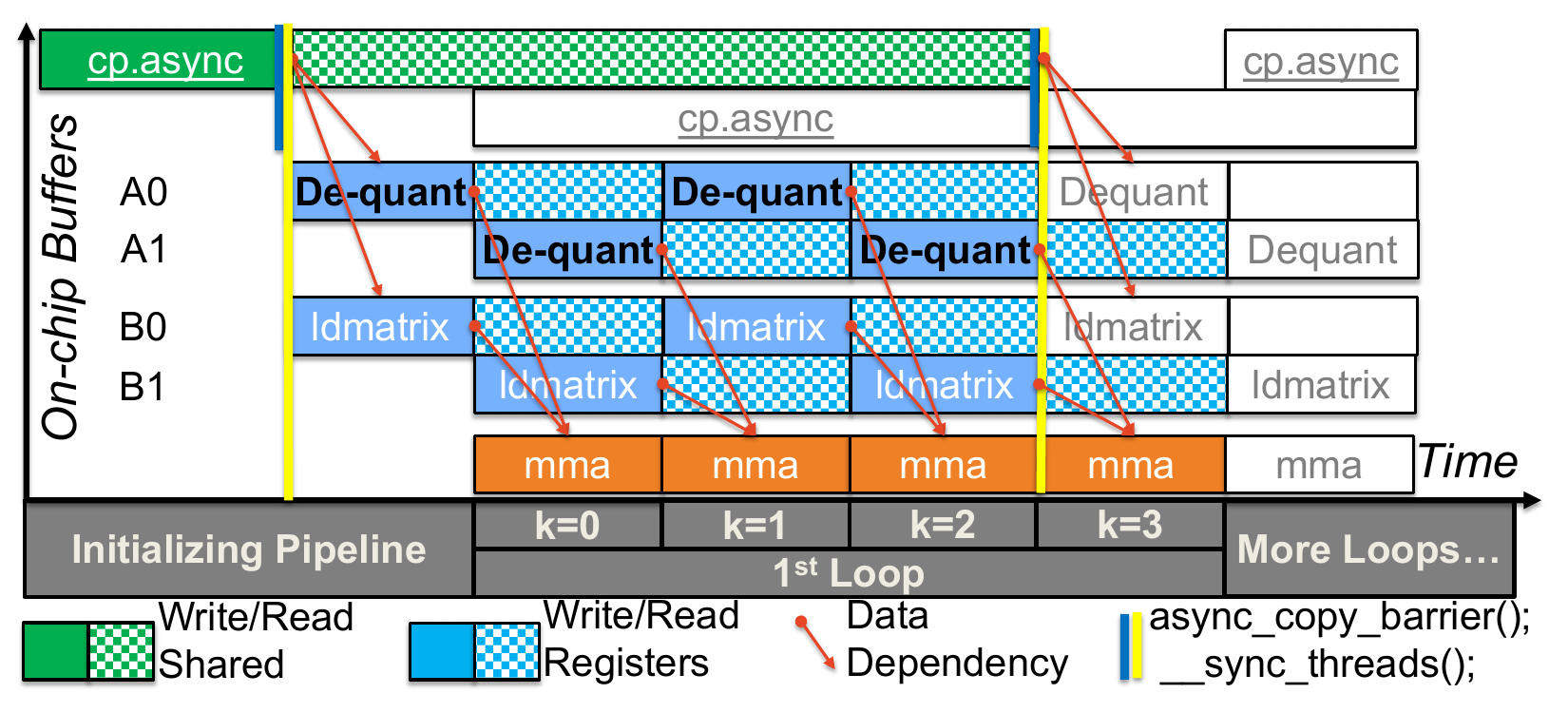}
    }
    \caption{Software Pipeline of \KERNEL{} GPU Kernel.}
    \label{fig:KernelDesign}
\end{figure}

\paragraph{Slice-by-slice De-quantization}
Instead of de-quantizing all the weights at once, we de-quantize the FPx weights slice by slice.
As shown in Figure \ref{fig:FourStageDequant}, we assume that an FPx weights tile and an FP16 activation tile are already copied from DRAM to \textit{shared memory}.
The whole tile of weight in \textit{shared memory} is then de-quantized in several steps.
In each step, only a slice of FPx weights is loaded from \textit{shared memory} to \textit{registers}, de-quantized into FP16 weights with \textit{SIMT-Efficient GPU Runtime} (Section \ref{sec:GPURuntime}), and then stored in the register buffer $A1$ or $A2$ as inputs for Tensor Cores.
$A_{Slice}$ and $B_{Slice}$ are then multiplied using Tensor Cores.

Compared to de-quantizing the whole tile at once, our slice-by-slice de-quantization reduces the number of registers required to store the FP16 weights by $4\times$, significantly reducing register pressure.
Besides, more opportunities are created for instruction-level parallelism, since Censor Cores can be used immediately for computations once a slice of weights is de-quantized, rather than waiting for the entire tile.

\paragraph{Effective Overlapping}
The software pipeline is illustrated via the space-time diagram in Figure \ref{fig:SpaceTimeDiagram}, where SIMT cores (working on de-quantization), Tensor Cores (working on matrix multiplication), and GPU memory hierarchy work collaboratively, achieving high instruction-level parallelism.

Firstly, global memory read is executed asynchronously using the \textit{cp.async}~\cite{Ampere_WhitePaper} intrinsic, fully overlapped with other operations.
Memory barrier and thread block synchronization are issued after the third slice is processed (at the end of k=2), making sure that the data for the next main loop is ready in \textit{shared memory} so that the \textit{"De-quant"} (de-quantization) and the \textit{"ldmatrix"} operations can be started when k=3.

Secondly, shared memory read is also overlapped with tensor core operations.
When the $i_{th}$ slice is being computed, the data of the $(i+1)_{th}$ slice are read from \textit{shared memory} simultaneously via \textit{"De-quant"} and \textit{"ldmatrix"}.

Last but not least, the SIMT core operations for weight de-quantization are also effectively overlapped with Tensor Core operations.
Within the \textit{"De-quant"} process of the $i_{th}$ slice, the FPx weights are first loaded from \textit{shared memory} to \textit{registers} using the hardware intrinsic \textit{load shared (LDS)}, and then immediately de-quantized into FP16 weights with SIMT cores.
At the same time, Tensor Cores are computing the $(i-1)_{th}$ slice with no data dependency.

\section{Implementation}
We implemented the \KERNEL{} kernel supporting matrix multiply $C = A \times B$, where $A$ is the weight matrix of shape $[M, K]$ and $B$ is the activation matrix of shape $[K, N]$.
The weight matrices are stored in our customized format described in Section \ref{sec:WeightPrepacking}, and the input and output activation matrices are stored in column-major.
Thus, our \KERNEL{} kernel could be a drop-in replacement of cuBLAS kernels in inference frameworks for quantized LLMs.
Our GPU kernel is implemented with more than 1.2K lines of CUDA codes, on top of the code of Flash-LLM~\cite{Flash-LLM-Github}.
Our \KERNEL{} kernels could be compiled separately into a $.so$ dynamic link-able library, and we provide a set of C++ APIs to call the kernels.
Thus, our kernels could be easily used and integrated.
Besides, we also provided C++ APIs to pre-pack the weight matrices (See Section \ref{sec:WeightPrepacking}).
More importantly, we provide new system support for end-to-end inference of quantized LLMs, by integrating our kernel into the state-of-the-art inference framework DeepSpeed~\cite{DeepSpeed}.
\section{Evaluation}

We evaluate the performance at two levels: kernel-level benchmarking using \KERNEL{} GPU kernels and model-level end-to-end inference using DeepSpeed integration (which we call \SYS{}).
The kernel-level evaluation is conducted on the NVIDIA A100-40GB platform with CUDA 11.8, and we mainly evaluate the performance of linear layers within LLMs during the token generation phase.
The utilization of each GPU hardware unit during runtime (Section \ref{sec:LinearLayerSpeedup_W8A16}) is measured using NVIDIA Nsight Compute\cite{NsightCompute}.
For end-to-end evaluations, we conduct the inference of typical LLMs on the NVIDIA A100-SXM4-80GB DGX platform with CUDA 11.8.
The inference latency and the latency breakdown (Section \ref{sec:e2e-inference}) are measured using NVIDIA Nsight System\cite{NsightSystem}.

\subsection{Linear Layer Speedups to 8-/16- bit}
\label{sec:LinearLayerSpeedup_W8A16}

\paragraph{Workloads.}
We evaluate the performance of \KERNEL{} on linear layers under different shapes, coming from the shapes of the weight matrices within LLaMA models~\cite{llama1} (llama-7b, llama-13b, llama-33b, and llama-65b) and OPT models~\cite{OPT-Models} (OPT-30b, OPT-65b, and OPT-175b).
For each model, we evaluated the latency of each GPU kernel at three typical inference batch sizes, i.e. 8, 16, and 32.

\paragraph{Baselines.}
The baselines we compare include the W16A16 kernels from cuBLAS~\cite{cuBLAS} and the W8A16 kernels from TensorRT-LLM (commit: 6837c81)~\cite{TensorRT-LLM}.
What's more, we also include the W4A16 (FP4) support from BitsandBytes (commit: f1ef74f)~\cite{BitsAndBytes} as a baseline.

\begin{figure*}
    \centering
    \includegraphics[width=0.98\textwidth]{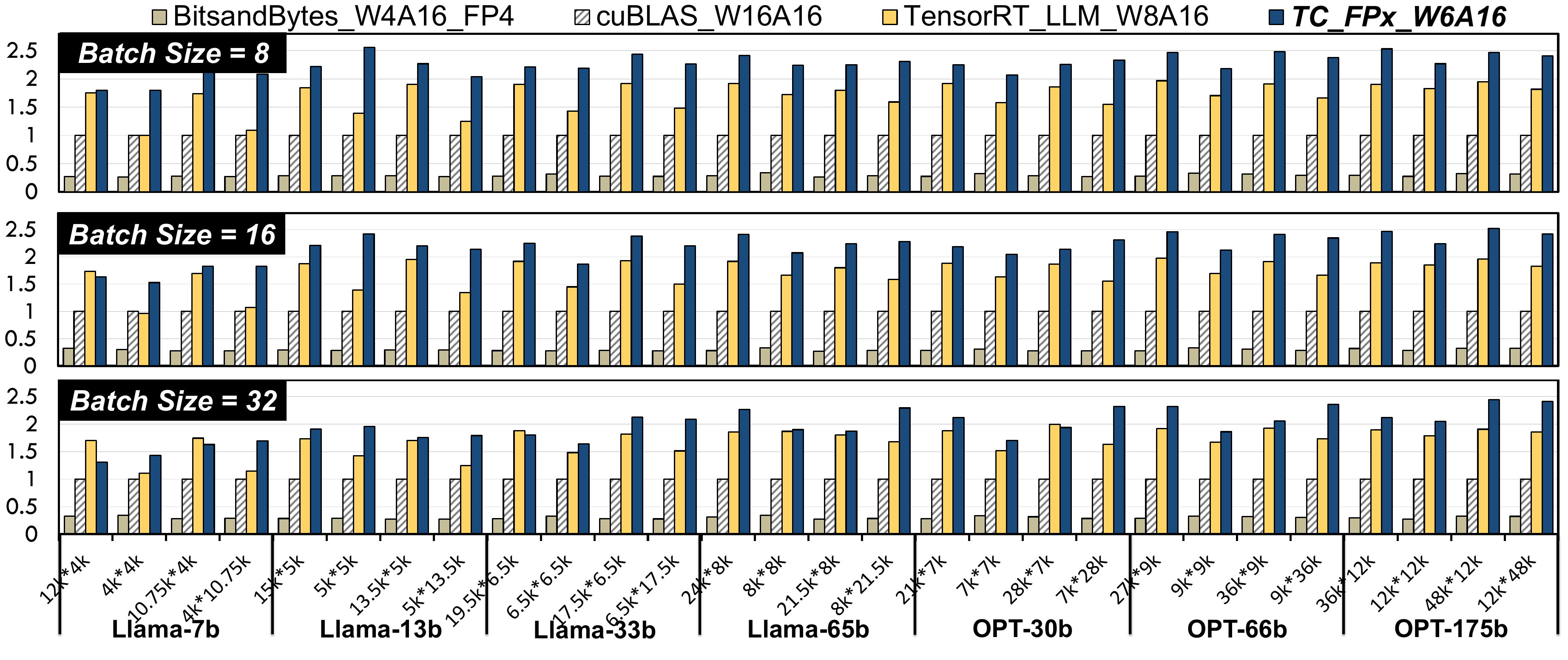}
    \caption{Linear layer speedups compared to the baselines for token generation phase.}
    \label{fig:LinearLayerSpeedup_W8A16}
\end{figure*}

\paragraph{Results.}
Figure \ref{fig:LinearLayerSpeedup_W8A16} shows the latency speedups of \KERNEL{} and other baselines.
We use the performance of cuBLAS to normalize the performance of all GPU kernels.
As shown in Figure \ref{fig:LinearLayerSpeedup_W8A16}, \KERNEL{} outperforms BitsandBytes (W4A16), cuBLAS (W16A16), and TensorRT\_LLM (W8A16, INT8 weights) by up to $8.9\times$, $2.6\times$, and $1.9\times$.
On average, \KERNEL{} outperforms BitsandBytes, cuBLAS and TensorRT\_LLM by $7.6\times$/$7.5\times$/$6.6\times$, $2.2\times$/$2.2\times$/$2.0\times$, and $1.3\times$/$1.3\times$/$1.2\times$ when the batch size is 8/16/32, respectively.

\begin{figure}
    \centering
    \subfloat[Tensor core and DRAM utilization at different batch sizes.
    \label{fig:KernelProfiling_TC}]{
        \includegraphics[width=0.9\linewidth]{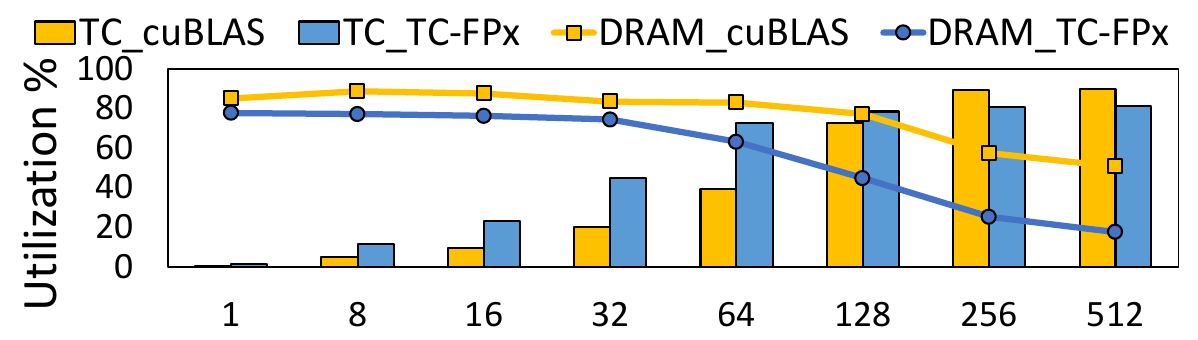}
    }
    
    \subfloat[ALU and FMA unit utilization at different batch sizes.
    \label{fig:KernelProfiling_SIMT}]{
        \includegraphics[width=0.9\linewidth]{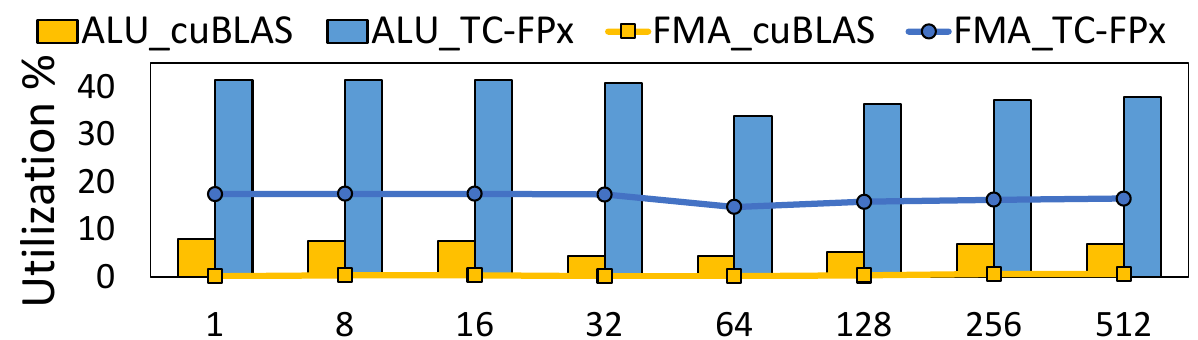}
    }
    \caption{Performance Analysis.}
    \label{fig:PerformanceAnalysis}
\end{figure}

\paragraph{Performance Analysis}
With extensive kernel profiling, We demonstrate the utilization\footnote{"Utilization" typically refers to the degree to which a particular hardware resource is being actively used during the execution of a GPU kernel.} of each GPU hardware unit and provide more in-depth insights into the source of our performance improvements.
During the execution of linear layers, as for the cuBLAS baseline, 
the DRAM bandwidth (shown as the yellow lines in Figure \ref{fig:KernelProfiling_TC}) is almost exhausted (>80\%) while the GPU Tensor Cores (shown as the yellow bar in Figure \ref{fig:KernelProfiling_TC}) are not fully used (<50\%), when the inference batch size is smaller than 128.
It is a common issue during the inference of large language models caused by the \textbf{auto-regressive inference} scheme of large language models.
With our support of 6-bit quantization, the DRAM access is significantly reduced (up to $2.7\times$), mitigating the bottleneck of insufficient DRAM bandwidth.
Consequently, the Tensor Cores can be more effectively utilized for matrix computations, shown as blue bars compared to yellow bars in Figure \ref{fig:KernelProfiling_TC}.
In summary, our kernel mitigated the "memory wall" issue and achieved higher computational efficiency (higher utilization of Tensor Cores) by supporting 6-bit quantization on Tensor Cores.

Furthermore, it explains that our kernel can outperform TensorRT-LLM's W8A16 kernel because we are more effective in reducing DRAM access of model weights.
Note that the performance of our \KERNEL{} kernel, cuBLAS kernel, and TensorRT-LLM's W8A16 kernel will eventually converge to the same performance when the inference batch size is larger (bigger than 128), as their performance will all be bounded by the peak computing power of Tensor Cores.

We also observed that BitsandBytes is constantly slower than cuBLAS, which is $29.6\%$ as fast as cuBLAS on average.
After further investigation, we found that BitsandBytes adopted the dual-kernel method (discussed in Section \ref{sec:choices}) to support FP4 quantization.
During the execution of the first kernel, the FP4 model weights will be first loaded from global memory, de-quantized into FP16, and then written back to global memory in the FP16 data type.
After that, a normal cuBLAS kernel is launched computing the matrix multiplication as the second kernel.
Thus, the FP4 GPU kernel is always slower than the original FP16 cuBLAS kernel due to the overhead of the extra GPU kernel for FP4 de-quantization.

\paragraph{Analysis of on-the-fly De-quantization}
Figure \ref{fig:KernelProfiling_SIMT} shows the overhead of FP6-to-FP16 de-quantization in two aspects.
On one hand, the FP6-to-FP16 de-quantization introduces a significant number of bit-wise operations even with our SIMT-efficient designs.
As a result, the utilization of the Arithmetic/Logic Unit (ALU) has increased from $6.36\%$ to $38.8\%$ on average.
It is also strong evidence that the SIMT-efficient designs (Section \ref{sec:GPURuntime}) for de-quantization are essential.
On the other hand, the FP6-to-FP16 de-quantization also introduces more float-point multiplications, computing the multiplication between the weights and the quantization scales. 
On average, the utilization of the FMA unit is increased from $0.33\%$ to $16.64\%$.
Given that both ALU and FMA units are part of the SIMT cores, the de-quantization operations will not consume the computing power of Tensor Cores.
More importantly, the runtime overhead of SIMT cores can be effectively hidden by overlapping these SIMT instructions with other operations, with our novel designs described in Section \ref{sec:PipelineDesign}.

\subsection{Performance Comparison to 4-bit}
\label{sec:LinearLayerSpeedup_W4A16}

\paragraph{Workloads}

As described in Section \ref{sec:BetterTradeoff}, 6-bit quantization is more appealing than 4-bit quantization in terms of preserving model quality.
However, we still compare the performance of our W6A16 kernels to the state-of-the-art W4A16 kernels, fully demonstrating that our 6-bit quantization can achieve comparable inference speed to the existing 4-bit quantization methods.
We evaluate the performance of the linear layers within the LLaMA-65b model~\cite{llama1} under different batch sizes.

\paragraph{Baselines}
The major baselines here include the W4A16 support of row-wise quantization (Coarse-grained\_W4A16) and the W4A16 support of group-wise quantization (Fine-grained\_W4A16) from TensorRT-LLM~\cite{TensorRT-LLM} (commit: 6837c81) with state-of-the-art performance.
We also include cuBLAS~\cite{cuBLAS} here as the performance baseline, clearly showing the benefits of each quantization method.

\begin{figure}
    \centering
    \includegraphics[width=1\columnwidth]{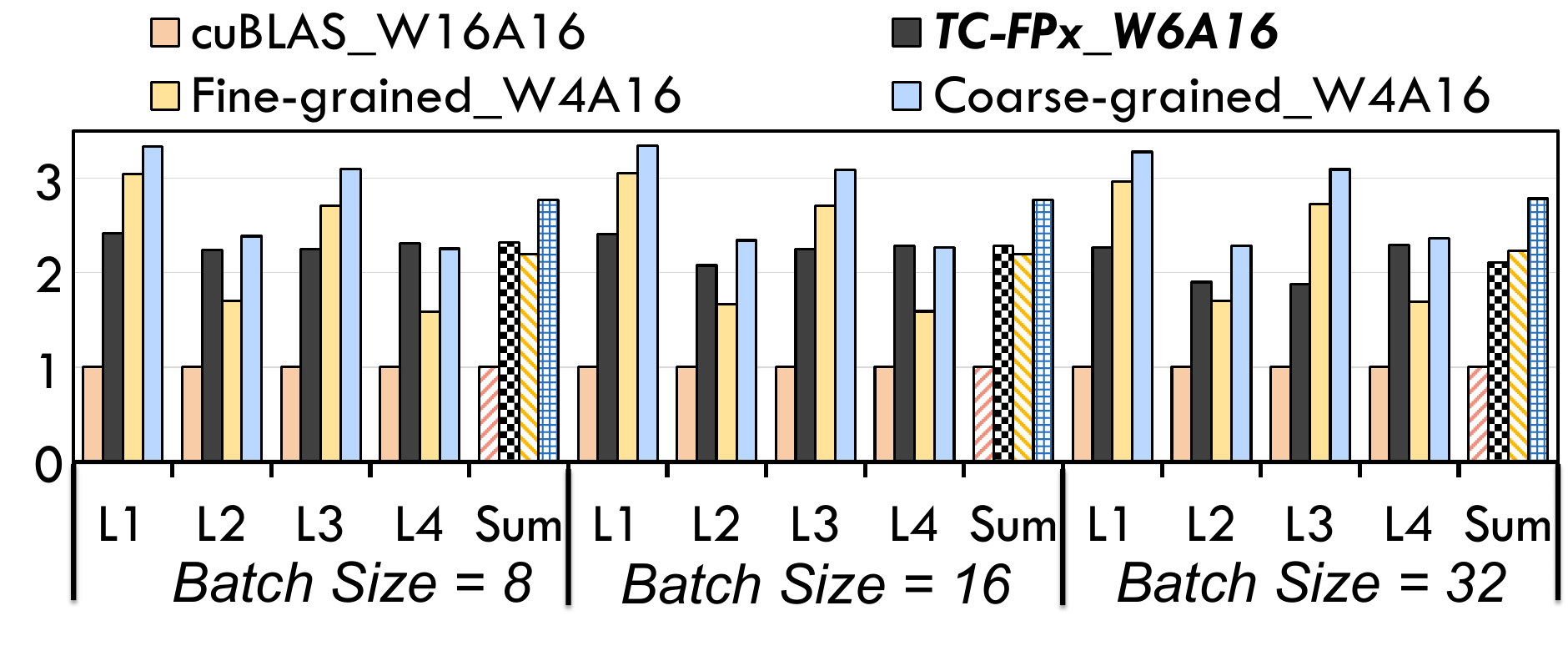}
    \caption{Linear layer speedups compared to using 4-bit weights for token generation phase of the LLaMA-65b model.}
    \label{fig:LinearLayerSpeedup_W4A16}
\end{figure}

\paragraph{Results}
Figure \ref{fig:LinearLayerSpeedup_W4A16} shows the latency speedups of \KERNEL{} and other baselines running four different linear layers (i.g. L1, L2, L3, and L4) within the LLaMA-65b models.
We use cuBLAS' performance to normalize the performance of other GPU kernels.
As shown in Figure \ref{fig:LinearLayerSpeedup_W4A16}, \KERNEL{}\_W6A16, Fine-grained\_W4A16, and Coarse-grained\_W4A16 outperform cuBLAS\_W16A16 by up to $2.4\times$, $3.0\times$, and $3.3\times$.
More importantly, \KERNEL{} achieves similar performance with Fine-grained\_W4A16, which is $1.06\times$/$1.04\times$/$0.94\times$ faster than Fine-grained\_W4A16 when running all these linear layers at batch size 8/16/32, respectively.
Besides, \KERNEL{} is only $16 \%$ / $17 \%$ / $24 \%$ slower than Coarse-grained\_W4A16 at batch size 8/16/32.
Since 6-bit quantization can provide significantly higher model quality, it is a worthwhile trade-off.

\subsection{End2End Inference}
\label{sec:e2e-inference}

\paragraph{Workloads}
We evaluate the end-to-end inference performance of \SYS{} on large language models of various model sizes, i.g. \underline{LLaMA-13b}~\cite{llama2}, \underline{OPT-30b}~\cite{OPT-Models}, and \underline{LLaMA-70b}~\cite{llama2}.
For each model, we evaluated its token generation throughput at different batch sizes, starting from 1 until GPU memory is exhausted.

\paragraph{Metric.}
We use the metric \textbf{tokens per GPU-second} to indicate the \textit{normalized inference throughput} with the consideration of both execution time and hardware cost (i.e., the number of GPUs used).
It is calculated with this equation:
\begin{equation}
    Inference\_Performance = \frac{N_{token}}{\sum_{i=1}^{N_{GPU}} T_i}
\end{equation}
$N_{token}$ means the number of tokens generated, whereas $N_{GPU}$ and $T_i$ mean the GPU number and the time spent on the i'th GPU for execution.
We use this metric to evaluate the end-to-end inference performance in this section.

\paragraph{Settings and Baselines}
We set the prefill/prompt length of each request to 0.5K, and generate 1.5K tokens for each request ignoring the \textit{"EOS"} (end of sequence) token.
We integrate our \KERNEL{} kernel into DeepSpeed~\cite{DeepSpeed} for end-to-end evaluation and call this new system support \SYS{}.
The baseline for comparison is the FP16 execution of the original DeepSpeed system.
With our \SYS{}, only a single 80GB A100 GPU is used for the inference for all the workloads, including the LLaMA-70b model~\cite{llama2}.
In contrast, two 80GB A100 GPUs are used for the inference of the LLaMA-70b model for the FP16 baseline, since the model weights ($\approx130$ GB) can not be fit into a single GPU.

\begin{figure}
\centering
    \subfloat[Generation throughput. \label{fig:GenerationThroughput-70b}]{
        \includegraphics[width=0.49\columnwidth]{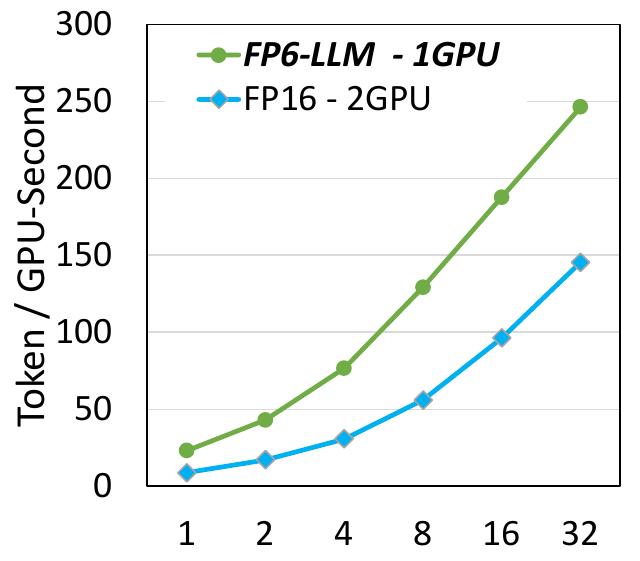}
    }
    \subfloat[Inference latency breakdown. \label{fig:LatencyBreakdown-70b}]{
        \includegraphics[width=0.49\columnwidth]{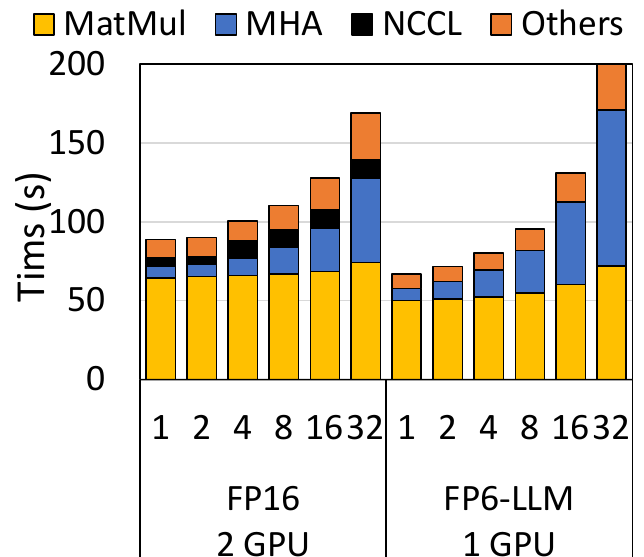}
    }
    \caption{LLaMA-70b inference at different batch sizes. \textit{MatMul}: linear layers, implemented with cuBLAS or our \KERNEL{}; \textit{MHA}: multi-head attention; \textit{NCCL}: cross-GPU communications; \textit{Others}: other GPU kernels or GPU idle time.}
    \label{fig:Inference-70b}
\end{figure}

\paragraph{LLaMA-70b}
Figure \ref{fig:GenerationThroughput-70b} shows token generation throughput on the LLaMA-70b model using our \SYS{} (\SYS{}-1GPU) and the FP16 baseline (FP16-2GPU).
According to our experiments, both our \SYS{} and FP16 baseline can at most set the inference batch size to 32 before running out of GPU memory, whereas \SYS{} only requires a single GPU and the baseline uses two GPUs.
The results show that \SYS{} can achieve $1.69\times$-$2.65\times$ higher normalized inference throughput than the FP16 baseline.

We conduct a careful latency breakdown of this end-to-end inference process.
As shown in Figure \ref{fig:LatencyBreakdown-70b}, our \KERNEL{} kernel (used in \SYS{}) is $1.20\times$ faster than cuBLAS kernel (used in FP16 baseline) on average, even \textbf{with half number of GPUs}.
Besides, the NCCL~\cite{NCCL} overhead (cross-GPU communications) is fully avoided using \SYS{} since only a single GPU is required.
We also notice that the FP16 baseline has accelerated the computation of MHA (multi-head attention) with 2-way tensor parallelism~\cite{TensorParallelism}.
Overall, our \SYS{} achieves up to $2.65\times$ higher throughput than the FP16 baseline as for \textit{tokens per GPU-second}.

\begin{figure}
\centering
    \subfloat[Generation throughput. \label{fig:GenerationThroughput-30b}]{
        \includegraphics[width=0.49\columnwidth]{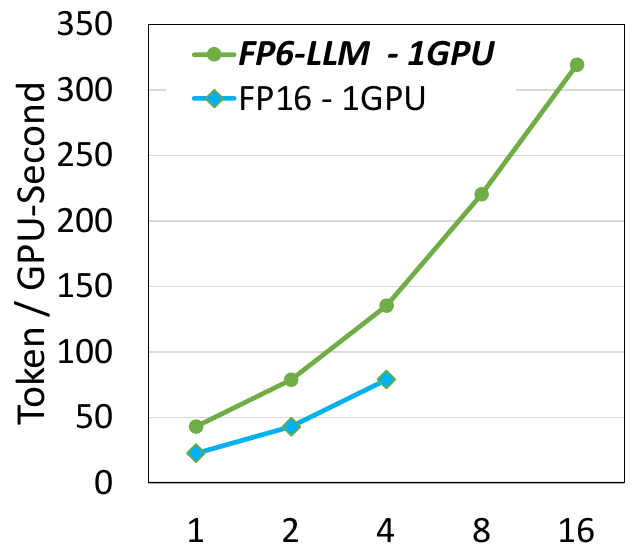}
    }
    \subfloat[Inference latency breakdown. \label{fig:LatencyBreakdown-30b}]{
        \includegraphics[width=0.49\columnwidth]{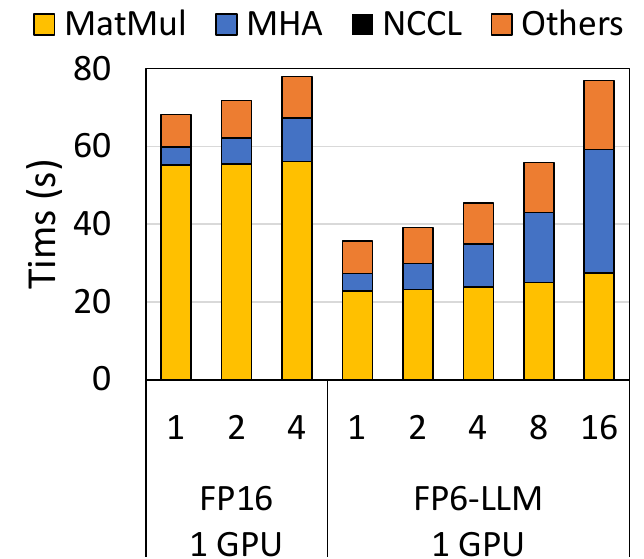}
    }
    \caption{OPT-30b inference at different batch sizes.}
    \label{fig:Inference-30b}
\end{figure}

\paragraph{OPT-30b}
Figure \ref{fig:GenerationThroughput-30b} shows token generation throughput on the OPT-30b model using \SYS{} (\SYS{}-1GPU) and the FP16 baseline (FP16-1GPU).
According to our experiments, \SYS{} can at most set the inference batch size to 16 before running out of GPU memory while the FP16 baseline can at most serve 4 requests in a batch.
As a result, \SYS{} can at most achieve 319.1 tokens per GPU-second ($4.05\times$ higher) with batch size 16 while the FP16 baseline can at most achieve 78.8 tokens per GPU-second with batch size 4, given the same GPU budget.
Besides, \SYS{} can achieve $1.91\times$/$1.84\times$/$1.72\times$ higher generation throughput compared to the FP16 baseline when their batch sizes are set to 1/2/4.
These overall performance improvements mainly come from the reduction of time in executing linear layers.
As shown in Figure \ref{fig:LatencyBreakdown-30b}, \KERNEL{} kernel is $2.39\times$ faster than the FP16 cuBLAS kernel on average.

\begin{figure}
\centering
    \subfloat[Generation throughput. \label{fig:GenerationThroughput-13b}]{
        \includegraphics[width=0.49\columnwidth]{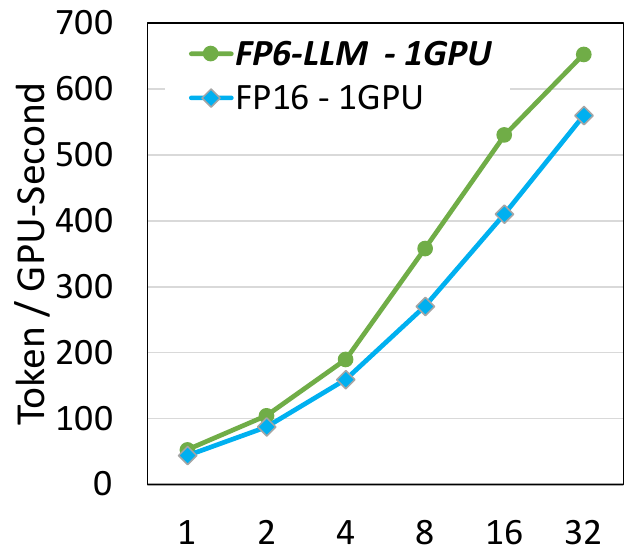}
    }
    \subfloat[Inference latency breakdown. \label{fig:LatencyBreakdown-13b}]{
        \includegraphics[width=0.49\columnwidth]{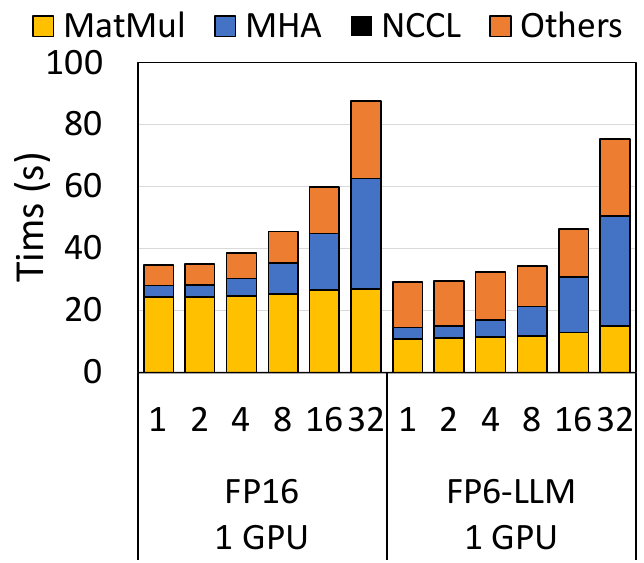}
    }
    \caption{LLaMA-13b inference at different batch sizes.}
    \label{fig:Inference-13b}
\end{figure}

\paragraph{LLaMA-13b}
Figure \ref{fig:GenerationThroughput-13b} shows the token generation throughput on the LLaMA-13b model using \SYS{} (\SYS{}-1GPU) and the FP16 baseline (FP16-1GPU).
According to the experiments, \SYS{} and the FP16 baseline can at most set the inference batch size to 32 before running out of memory.
On average, \SYS{} can achieve $1.23\times$ higher generation throughput compared to the FP16 baseline using the same batch size.
The overall performance improvements on this model are less significant compared to the previous two models due to the \textit{non-kernel overhead}.
According to Figure \ref{fig:LatencyBreakdown-13b}, the execution time of linear layers has been significantly reduced ($2.11\times$ faster on average) with \KERNEL{} kernel.
However, the portion of running other GPU kernels plus the GPU idle time increases, weakening the overall performance gains.
The reason is that GPUs tend to have a larger proportion of idle time due to kernel launch latency and GPU synchronizations as the model size gets smaller.

\section{Related Work}

\paragraph{Six-bit Quantization}
\cite{zeroquant42} shows that FP6 performs robustly across various algorithms and tasks, demonstrating its superiority in accuracy and versatility.
Besides,~\cite{microscaling} verified that the FP6 data format can closely match FP32 for inference after quantization-aware fine-tuning.
However, there is no hardware support for the proposed data types.
Their inference/training experiments can only be done via software emulations.
\SYS{} can provide high-performance GPU support for the inference of LLMs after FP6 quantization.

\paragraph{System Supports for Quantization}
\underline{\textit{TensorRT-LLM}}~\cite{TensorRT-LLM} has state-of-the-art kernel supports for weight-only quantization.
However, it only supports weights in INT4 (W4A16 ~\cite{OPTQ,awq}) or INT8 (W8A16 and W8A8~\cite{smoothquant}) data types while we provide better trade-offs by supporting weights in 6 bits.
Besides, TensorRT-LLM does not support float-point data type (e.g. FP6), which is much more complicated to de-quantize during runtime than the integer type.
\underline{\textit{Bitsandbytes}}~\cite{BitsAndBytes} mainly supports INT8 weights (W8A8) and has very naive support for FP4 (W4A16) with poor performance.
\underline{\textit{Llama.cpp}}~\cite{llama.cpp} has 2-bit, 3-bit, 4-bit, 5-bit, 6-bit, and 8-bit quantization support on CPUs or GPU SIMT cores.
However, it does not support weights in float point data type and it can not make use of GPU tensor cores.
\underline{AWQ}~\cite{awq} has GPU kernel implementation~\cite{awq-github} for memory-efficient 4-bit Linear (W4A16) in PyTorch.
\underline{OPTQ}~\cite{OPTQ} has a basic GEMV implementation~\cite{OPTQ-github} for weights in INT3 (W3A16) data type.
To the best of our knowledge, this work is the first system supporting weight-only quantization with FP6 weights on Tensor cores.

\paragraph{Related Design Techniques}
\cite{zeroquantv2} and~\cite{zeroquant42} has previously proposed 4+2 weight split.
However, they only presented intuitive thoughts without comprehensive system designs.
Flash-LLM~\cite{Flash-LLM} has proposed the load-as-sparse and compute-as-dense approach for the weight-only-pruning. It does not tackle the problems of supporting quantization.

\section{Conclusions}
In this paper, we introduce \KERNEL{}, the first full-stack GPU kernel design scheme with unified tensor core support for float-point weights of various quantization bit-width, mitigating the "memory wall" issues during LLM inference.
We integrate \KERNEL{} kernel into a state-of-the-art inference system, providing new end-to-end support (called \SYS{}) for quantized LLM inference, where better trade-offs between inference cost and model quality are achieved.
\SYS{} tackles the problems of hardware-unfriendly memory access and high computation overhead of de-quantization with a set of novel techniques, achieving faster inference speed with significantly less GPU memory.
Evaluations show that \SYS{} enables the inference of LLaMA-70b using only a single GPU, achieving $1.69\times$-$2.65\times$ higher normalized inference throughput than the FP16 baseline.
Besides, \SYS{} improves the inference throughput of OPT-30b by $1.72\times$-$4.05\times$.




\bibliographystyle{plain}
\bibliography{references}

\begin{thebibliography}{10}

\bibitem{GPT3}
Tom~B. Brown, Benjamin Mann, Nick Ryder, Melanie Subbiah, Jared Kaplan, Prafulla Dhariwal, Arvind Neelakantan, Pranav Shyam, Girish Sastry, Amanda Askell, Sandhini Agarwal, Ariel Herbert-Voss, Gretchen Krueger, Tom Henighan, Rewon Child, Aditya Ramesh, Daniel~M. Ziegler, Jeffrey Wu, Clemens Winter, Christopher Hesse, Mark Chen, Eric Sigler, Mateusz Litwin, Scott Gray, Benjamin Chess, Jack Clark, Christopher Berner, Sam McCandlish, Alec Radford, Ilya Sutskever, and Dario Amodei.
\newblock Language models are few-shot learners, 2020.

\bibitem{chen2021evaluating}
Mark Chen, Jerry Tworek, Heewoo Jun, Qiming Yuan, Henrique Ponde de~Oliveira Pinto, Jared Kaplan, Harri Edwards, Yuri Burda, Nicholas Joseph, Greg Brockman, et~al.
\newblock Evaluating large language models trained on code.
\newblock {\em arXiv preprint arXiv:2107.03374}, 2021.

\bibitem{BitsAndBytes}
Tim Dettmers.
\newblock bitsandbytes.
\newblock \url{"https://github.com/TimDettmers/bitsandbytes"}, 2023.

\bibitem{LLM.INT8}
Tim Dettmers, Mike Lewis, Younes Belkada, and Luke Zettlemoyer.
\newblock Llm.int8(): 8-bit matrix multiplication for transformers at scale, 2022.

\bibitem{OPTQ-github}
Elias Frantar, Saleh Ashkboos, Torsten Hoefler, and Dan Alistarh.
\newblock gptq.
\newblock \url{"https://github.com/IST-DASLab/gptq"}, 2022.

\bibitem{frantar2022gptq}
Elias Frantar, Saleh Ashkboos, Torsten Hoefler, and Dan Alistarh.
\newblock Gptq: Accurate post-training quantization for generative pre-trained transformers.
\newblock {\em arXiv preprint arXiv:2210.17323}, 2022.

\bibitem{OPTQ}
Elias Frantar, Saleh Ashkboos, Torsten Hoefler, and Dan Alistarh.
\newblock Optq: Accurate quantization for generative pre-trained transformers.
\newblock In {\em The Eleventh International Conference on Learning Representations}, 2022.

\bibitem{llama.cpp}
Georgi Gerganov.
\newblock llama.cpp.
\newblock \url{"https://github.com/ggerganov/llama.cpp"}, 2023.

\bibitem{Copilot}
Github.
\newblock Copilot.
\newblock \url{"https://github.com/features/copilot"}, 2022.

\bibitem{Bard}
Google.
\newblock Bard.
\newblock \url{"https://bard.google.com/"}, 2023.

\bibitem{IEEE754}
William Kahan.
\newblock Ieee standard 754 for binary floating-point arithmetic.
\newblock {\em Lecture Notes on the Status of IEEE}, 754(94720-1776):11, 1996.

\bibitem{FullStackOptimization}
Sehoon Kim, Coleman Hooper, Thanakul Wattanawong, Minwoo Kang, Ruohan Yan, Hasan Genc, Grace Dinh, Qijing Huang, Kurt Keutzer, Michael~W Mahoney, et~al.
\newblock Full stack optimization of transformer inference: a survey.
\newblock {\em arXiv preprint arXiv:2302.14017}, 2023.

\bibitem{li2023starcoder}
Raymond Li, Loubna~Ben Allal, Yangtian Zi, Niklas Muennighoff, Denis Kocetkov, Chenghao Mou, Marc Marone, Christopher Akiki, Jia Li, Jenny Chim, Qian Liu, Evgenii Zheltonozhskii, Terry~Yue Zhuo, Thomas Wang, Olivier Dehaene, Mishig Davaadorj, Joel Lamy-Poirier, João Monteiro, Oleh Shliazhko, Nicolas Gontier, Nicholas Meade, Armel Zebaze, Ming-Ho Yee, Logesh~Kumar Umapathi, Jian Zhu, Benjamin Lipkin, Muhtasham Oblokulov, Zhiruo Wang, Rudra Murthy, Jason Stillerman, Siva~Sankalp Patel, Dmitry Abulkhanov, Marco Zocca, Manan Dey, Zhihan Zhang, Nour Fahmy, Urvashi Bhattacharyya, Wenhao Yu, Swayam Singh, Sasha Luccioni, Paulo Villegas, Maxim Kunakov, Fedor Zhdanov, Manuel Romero, Tony Lee, Nadav Timor, Jennifer Ding, Claire Schlesinger, Hailey Schoelkopf, Jan Ebert, Tri Dao, Mayank Mishra, Alex Gu, Jennifer Robinson, Carolyn~Jane Anderson, Brendan Dolan-Gavitt, Danish Contractor, Siva Reddy, Daniel Fried, Dzmitry Bahdanau, Yacine Jernite, Carlos~Muñoz Ferrandis, Sean Hughes, Thomas Wolf, Arjun Guha, Leandro von
  Werra, and Harm de~Vries.
\newblock Starcoder: may the source be with you!, 2023.

\bibitem{awq}
Ji~Lin, Jiaming Tang, Haotian Tang, Shang Yang, Xingyu Dang, Chuang Gan, and Song Han.
\newblock Awq: Activation-aware weight quantization for llm compression and acceleration, 2023.

\bibitem{awq-github}
Ji~Lin, Jiaming Tang, Haotian Tang, Shang Yang, Xingyu Dang, and Song Han.
\newblock llm-awq.
\newblock \url{"https://github.com/mit-han-lab/llm-awq"}, 2023.

\bibitem{mftcoder2023}
Bingchang Liu, Chaoyu Chen, Cong Liao, Zi~Gong, Huan Wang, Zhichao Lei, Ming Liang, Dajun Chen, Min Shen, Hailian Zhou, Hang Yu, and Jianguo Li.
\newblock Mftcoder: Boosting code llms with multitask fine-tuning.
\newblock {\em arXiv preprint arXiv}, 2023.

\bibitem{marcinkiewicz1994building}
Mary~Ann Marcinkiewicz.
\newblock Building a large annotated corpus of english: The penn treebank.
\newblock {\em Using Large Corpora}, page 273, 1994.

\bibitem{merity2016pointer}
Stephen Merity, Caiming Xiong, James Bradbury, and Richard Socher.
\newblock Pointer sentinel mixture models.
\newblock In {\em International Conference on Learning Representations}, 2017.

\bibitem{DeepSpeed}
Microsoft.
\newblock Deepspeed github.
\newblock \url{"https://github.com/microsoft/DeepSpeed"}, 2023.

\bibitem{Ampere_WhitePaper}
NVIDIA.
\newblock Nvidia a100 tensor core gpu architecture.
\newblock \url{"https://images.nvidia.com/aem-dam/en-zz/Solutions/data-center/nvidia-ampere-architecture-whitepaper.pdf"}, 2020.

\bibitem{Hopper_WhitePaper}
NVIDIA.
\newblock Nvidia h100 tensor core gpu architecture.
\newblock \url{"https://www.hpctech.co.jp/catalog/gtc22-whitepaper-hopper_v1.01.pdf"}, 2022.

\bibitem{cuBLAS}
NVIDIA.
\newblock cublas.
\newblock \url{"https://developer.nvidia.com/cublas"}, 2023.

\bibitem{NsightCompute}
NVIDIA.
\newblock Nsight compute profiling guide.
\newblock \url{"https://docs.nvidia.com/nsight-compute/ProfilingGuide/#introduction"}, 2023.

\bibitem{NsightSystem}
NVIDIA.
\newblock Nsight system.
\newblock \url{"https://developer.nvidia.com/nsight-systems"}, 2023.

\bibitem{NCCL}
NVIDIA.
\newblock Nvidia collective communications library (nccl).
\newblock \url{"https://developer.nvidia.com/nccl"}, 2023.

\bibitem{TensorRT-LLM}
NVIDIA.
\newblock Tensorrt-llm.
\newblock \url{"https://github.com/NVIDIA/TensorRT-LLM/"}, 2023.

\bibitem{ChatGPT}
OpenAI.
\newblock Chatgpt.
\newblock \url{"https://openai.com/blog/chatgpt"}, 2022.

\bibitem{GPT4}
OpenAI.
\newblock Gpt-4 technical report, 2023.

\bibitem{raffel2020exploring}
Colin Raffel, Noam Shazeer, Adam Roberts, Katherine Lee, Sharan Narang, Michael Matena, Yanqi Zhou, Wei Li, and Peter~J Liu.
\newblock Exploring the limits of transfer learning with a unified text-to-text transformer.
\newblock {\em The Journal of Machine Learning Research}, 21(1):5485--5551, 2020.

\bibitem{microscaling}
Bita~Darvish Rouhani, Ritchie Zhao, Ankit More, Mathew Hall, Alireza Khodamoradi, Summer Deng, Dhruv Choudhary, Marius Cornea, Eric Dellinger, Kristof Denolf, Stosic Dusan, Venmugil Elango, Maximilian Golub, Alexander Heinecke, Phil James-Roxby, Dharmesh Jani, Gaurav Kolhe, Martin Langhammer, Ada Li, Levi Melnick, Maral Mesmakhosroshahi, Andres Rodriguez, Michael Schulte, Rasoul Shafipour, Lei Shao, Michael Siu, Pradeep Dubey, Paulius Micikevicius, Maxim Naumov, Colin Verrilli, Ralph Wittig, Doug Burger, and Eric Chung.
\newblock Microscaling data formats for deep learning, 2023.

\bibitem{TensorParallelism}
Mohammad Shoeybi, Mostofa Patwary, Raul Puri, Patrick LeGresley, Jared Casper, and Bryan Catanzaro.
\newblock Megatron-lm: Training multi-billion parameter language models using model parallelism.
\newblock {\em arXiv preprint arXiv:1909.08053}, 2019.

\bibitem{llama1}
Hugo Touvron, Thibaut Lavril, Gautier Izacard, Xavier Martinet, Marie-Anne Lachaux, Timothée Lacroix, Baptiste Rozière, Naman Goyal, Eric Hambro, Faisal Azhar, Aurelien Rodriguez, Armand Joulin, Edouard Grave, and Guillaume Lample.
\newblock Llama: Open and efficient foundation language models, 2023.

\bibitem{llama2}
Hugo Touvron, Louis Martin, Kevin Stone, Peter Albert, Amjad Almahairi, Yasmine Babaei, Nikolay Bashlykov, Soumya Batra, Prajjwal Bhargava, Shruti Bhosale, Dan Bikel, Lukas Blecher, Cristian~Canton Ferrer, Moya Chen, Guillem Cucurull, David Esiobu, Jude Fernandes, Jeremy Fu, Wenyin Fu, Brian Fuller, Cynthia Gao, Vedanuj Goswami, Naman Goyal, Anthony Hartshorn, Saghar Hosseini, Rui Hou, Hakan Inan, Marcin Kardas, Viktor Kerkez, Madian Khabsa, Isabel Kloumann, Artem Korenev, Punit~Singh Koura, Marie-Anne Lachaux, Thibaut Lavril, Jenya Lee, Diana Liskovich, Yinghai Lu, Yuning Mao, Xavier Martinet, Todor Mihaylov, Pushkar Mishra, Igor Molybog, Yixin Nie, Andrew Poulton, Jeremy Reizenstein, Rashi Rungta, Kalyan Saladi, Alan Schelten, Ruan Silva, Eric~Michael Smith, Ranjan Subramanian, Xiaoqing~Ellen Tan, Binh Tang, Ross Taylor, Adina Williams, Jian~Xiang Kuan, Puxin Xu, Zheng Yan, Iliyan Zarov, Yuchen Zhang, Angela Fan, Melanie Kambadur, Sharan Narang, Aurelien Rodriguez, Robert Stojnic, Sergey Edunov, and Thomas
  Scialom.
\newblock Llama 2: Open foundation and fine-tuned chat models, 2023.

\bibitem{AttentionIsAllYouNeed}
Ashish Vaswani, Noam Shazeer, Niki Parmar, Jakob Uszkoreit, Llion Jones, Aidan~N Gomez, {\L}ukasz Kaiser, and Illia Polosukhin.
\newblock Attention is all you need.
\newblock {\em Advances in neural information processing systems}, 30, 2017.

\bibitem{zeroquant42}
Xiaoxia Wu, Haojun Xia, Stephen Youn, Zhen Zheng, Shiyang Chen, Arash Bakhtiari, Michael Wyatt, Yuxiong He, Olatunji Ruwase, Leon Song, and Zhewei Yao.
\newblock Zeroquant(4+2): Redefining llms quantization with a new fp6-centric strategy for diverse generative tasks.
\newblock {\em arXiv preprint arXiv: 2312.08583}, 2023.

\bibitem{Flash-LLM}
Haojun Xia, Zhen Zheng, Yuchao Li, Donglin Zhuang, Zhongzhu Zhou, Xiafei Qiu, Yong Li, Wei Lin, and Shuaiwen~Leon Song.
\newblock Flash-llm: Enabling cost-effective and highly-efficient large generative model inference with unstructured sparsity.
\newblock {\em Proc. VLDB Endow.}, 17(2):211–224, oct 2023.

\bibitem{Flash-LLM-Github}
Haojun Xia, Zhen Zheng, Yuchao Li, Donglin Zhuang, Zhongzhu Zhou, Xiafei Qiu, Yong Li, Wei Lin, and Shuaiwen~Leon Song.
\newblock Flash-llm github.
\newblock \url{"https://github.com/AlibabaResearch/flash-llm"}, 2023.

\bibitem{smoothquant}
Guangxuan Xiao, Ji~Lin, Mickael Seznec, Hao Wu, Julien Demouth, and Song Han.
\newblock Smoothquant: Accurate and efficient post-training quantization for large language models, 2023.

\bibitem{yao2023zeroquant}
Zhewei Yao, Xiaoxia Wu, Cheng Li, Stephen Youn, and Yuxiong He.
\newblock Zeroquant-v2: Exploring post-training quantization in llms from comprehensive study to low rank compensation.
\newblock {\em arXiv preprint arXiv:2303.08302}, 2023.

\bibitem{zeroquantv2}
Zhewei Yao, Xiaoxia Wu, Cheng Li, Stephen Youn, and Yuxiong He.
\newblock Zeroquant-v2: Exploring post-training quantization in llms from comprehensive study to low rank compensation, 2023.

\bibitem{OPT-Models}
Susan Zhang, Stephen Roller, Naman Goyal, Mikel Artetxe, Moya Chen, Shuohui Chen, Christopher Dewan, Mona Diab, Xian Li, Xi~Victoria Lin, Todor Mihaylov, Myle Ott, Sam Shleifer, Kurt Shuster, Daniel Simig, Punit~Singh Koura, Anjali Sridhar, Tianlu Wang, and Luke Zettlemoyer.
\newblock Opt: Open pre-trained transformer language models, 2022.

\bibitem{ATOM}
Yilong Zhao, Chien-Yu Lin, Kan Zhu, Zihao Ye, Lequn Chen, Size Zheng, Luis Ceze, Arvind Krishnamurthy, Tianqi Chen, and Baris Kasikci.
\newblock Atom: Low-bit quantization for efficient and accurate llm serving, 2023.

\bibitem{zheng2023codegeex}
Qinkai Zheng, Xiao Xia, Xu~Zou, Yuxiao Dong, Shan Wang, Yufei Xue, Zihan Wang, Lei Shen, Andi Wang, Yang Li, Teng Su, Zhilin Yang, and Jie Tang.
\newblock Codegeex: A pre-trained model for code generation with multilingual evaluations on humaneval-x.
\newblock In {\em KDD}, 2023.

\bibitem{zhu2023survey}
Xunyu Zhu, Jian Li, Yong Liu, Can Ma, and Weiping Wang.
\newblock A survey on model compression for large language models.
\newblock {\em arXiv preprint arXiv:2308.07633}, 2023.

\end{thebibliography}

\end{document}